\documentclass[twoside,11pt]{article}
\usepackage{jmlr2e}
\usepackage{amsmath,amssymb}
\usepackage{graphicx,booktabs,multirow}
\newsavebox{\tablebox}
\newcommand{\fitwidth}[1]{\sbox{\tablebox}{#1}\ifdim\wd\tablebox>\textwidth\resizebox{\textwidth}{!}{\usebox{\tablebox}}\else\usebox{\tablebox}\fi}

\newcommand{\BoverNdOne}{0.07}

\newcommand{\FitTimeBig}{10 minutes}

\newcommand{\MemPeakDEight}{521\,MB}
\newcommand{\SlopeSummary}{from $-1.05$ to $-0.89$ across $d$}

\newcommand{\SpeedFitRange}{12--185$\times$}

\newcommand{\SpeedPredRange}{221--269$\times$}
\newcommand{\SweepRange}{6--11}

\newcommand{\UnsafeSweepsDOne}{353}

\newcommand{\SynStreamCells}{18}
\newcommand{\SynStreamBeatsKDE}{18}
\newcommand{\SynKDEOverStreamRange}{1.3--16}
\newcommand{\SynStreamOverOracleMed}{0.94}
\newcommand{\SynStreamOverOracleRange}{0.53--3.26}
\newcommand{\SynFcErrMed}{22}
\newcommand{\SynFcErrRange}{7--43}

\newcommand{\TraceTOneErrLow}{57}
\newcommand{\TraceTOneErrHigh}{6}
\newcommand{\CpuSpeedupRange}{4.4--5.1}
\newcommand{\GpuSpeedupRange}{1.3--4.2}

\newcommand{\PeriodRange}{23.66--24.21}
\newcommand{\PeriodMean}{23.97}
\newcommand{\PeriodTime}{240}

\newcommand{\FixedIseRatioRange}{0.999--1}
\newcommand{\FixedSpeedTwentyK}{1.02--1.04}

\newcommand{\FixedSpeedFiftyK}{1.16--1.4}

\newcommand{\FixedEpochs}{9--11}
\newcommand{\FixedSweeps}{9--10}

\newcommand{\DayNtest}{2{,}519}
\newcommand{\DayNdays}{105}
\newcommand{\DayStreamFcLagWeek}{0.004}
\newcommand{\DayAddWeekBLMLstream}{+0.010}

\newcommand{\DayStreamHL}{-3.393}
\newcommand{\DayStreamVsCondKNNHL}{0.31}

\newcommand{\DayStreamVsCondKDEHL}{0.39}

\newcommand{\DayStreamVsMDNHLAbs}{0.03}

\newcommand{\DayStreamVsTabPFNHLAbs}{0.06}

\newcommand{\DayStreamVsCondKNNHLL}{0.27}

\newcommand{\DayStreamVsCondKDEHLL}{0.47}

\newcommand{\DayStreamVsTabPFNHLLAbs}{0.02}

\newcommand{\NsWinsB}{4}
\newcommand{\NsCells}{6}
\newcommand{\NsConv}{0}
\newcommand{\NsRuns}{48}

\jmlrheading{}{2026}{}{}{}{}{Rahul Agarwal and Ankit Agarwal}
\ShortHeadings{Band-Limited Maximum Likelihood in High Dimensions}{Agarwal and Agarwal}
\firstpageno{1}

\begin{document}

\title{Taking Band-Limited Maximum Likelihood to High Dimensions: Linear Memory, the Parametric Rate, and
a Self-Tuning Cut-off}

\author{\name Rahul Agarwal \email rahul.jhu@gmail.com \\
       \addr Independent researcher (Ph.D., Biomedical Engineering, Johns Hopkins University)\\ Sammamish, WA, USA
       \AND
       \name Ankit Agarwal \email ankit.iiith@gmail.com \\
       \addr Independent researcher (M.B.A., University of Southern California)\\ San Jose, CA, USA}

\editor{}

\maketitle

\begin{abstract}
Band-limited maximum likelihood (BLML) estimates a density as the square of a band-limited function. The estimate is
non-negative, and when the square root of the true density is band-limited its error falls at the parametric rate
$n^{-1}$. Its algorithm bins the data, which is fast only in one or two dimensions: from three on every point has
its own bin, the kernel system is dense, and memory grows as $n^2$, 40\,GB at $10^5$ points. Whether the rate
survives in higher dimension, where kernel estimators slow down, had not been tested. We study
dense data, densities that fill their space, not data near a low-dimensional manifold. We show that the binned
equations are the stationarity condition of a strictly convex objective and minimise it by block coordinate descent
without forming the kernel matrix: memory is linear in $n$, a sweep costs $O(n^2d)$ and a fit takes about ten, and
one A100 GPU fits $10^6$ points in eight dimensions in about ten minutes. On band-limited truths the error then decays
as $n^{-1}$ in every dimension we tested, up to eight, while the slopes of kernel estimators flatten from $-0.67$ to
$-0.28$, and it is 2 to 18 times lower than a neural spline flow's. A streaming fit learns the cut-off, BLML's only
tuning parameter, from analytic gradients of the held-out likelihood, for about the cost of one more fit. On real
data BLML is the strongest nonparametric estimator we tested: it leads kernel and nearest-neighbour estimators on
electricity prices by 0.27 to 0.47 nats per hour, and a tuned kernel estimator on the full POWER and GAS benchmarks
by 0.39 and 1.58 nats per point, placing it between early and recent normalizing flows. It loses on densities with a
kink.
\end{abstract}


\section{Introduction}
\label{sec:intro}

Can a nonparametric density estimator keep the parametric rate as the dimension grows, and choose its own smoothing
as it goes? Kernel density estimators cannot: their mean integrated squared error (MISE) decays as
$n^{-4/(4+d)}$ \citep{wand1995,tsybakov2009}, and every added dimension slows them further. Higher-order and sinc
kernels buy a faster rate for smooth densities by returning estimates that go negative \citep{davis1975,hall1988order}.

\begin{figure}[t]
  \centering
  \includegraphics[width=\textwidth]{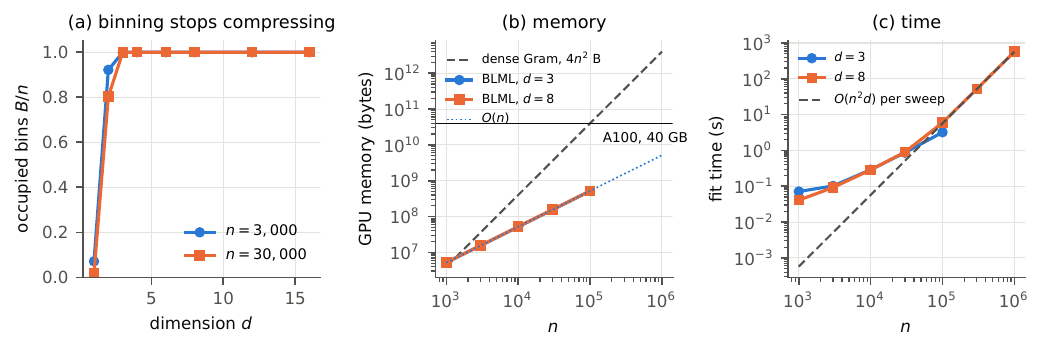}
  \caption{What makes high-dimensional BLML feasible. (a)~Binning compresses the data only for $d\le2$; from $d=3$
  on every point has its own bin, so a solver must handle $B=n$. (b)~Peak GPU memory of our solver grows as $O(n)$
  (dotted: linear fit), where a dense float32 Gram matrix grows as $4n^2$ bytes and exceeds an A100 beyond
  $n\approx10^5$. (c)~Fit time against $n$; dashed: the $O(n^2d)$ cost per sweep, anchored at the largest fit.}
  \label{fig:complexity}
\end{figure}

Band-limited maximum likelihood (BLML; \citealp{agarwal2017blml}) suggests a different answer. It assumes that
$f=g^2$ with $g$ band-limited to a cut-off $f_c$ and computes the exact maximum likelihood estimate in that class, a
squared kernel expansion that is non-negative by construction. In one dimension its error falls as $n^{-1}$. Its fast algorithm bins the
data, which makes it fast in one or two dimensions. From three dimensions on, binning no longer shrinks the problem
(Fig.~\ref{fig:complexity}a): every point carries its own coefficient, the kernel system is dense, and storing it takes
memory growing as $n^2$, 40\,GB at $10^5$ points (Fig.~\ref{fig:complexity}b). High-dimensional BLML was limited to small samples,
and so was the question of whether its rate survives there, where kernel estimators slow down. And the cut-off, the estimator's only tuning parameter, had to be chosen by refitting
the estimator tens of times.

We remove both obstacles and then ask how far the estimator goes.

\emph{A solver for high dimension} (Section~\ref{sec:method}). The binned BLML equations are the stationarity
condition of a strictly convex objective. We minimise it by block coordinate descent with exact one-dimensional
kernel tables and a cached matrix--vector product, never forming the Gram matrix: memory is linear in $n$
(\MemPeakDEight\ at $d=8$, $n=10^5$), a fit takes \SweepRange\ sweeps in every dimension, and one A100 GPU fits $10^6$
points in eight dimensions in about ten minutes.

\emph{The rate, measured} (Sections~\ref{sec:rate}--\ref{sec:experiments}). On band-limited truths BLML's MISE decays
as $n^{-1}$ in every dimension we tested, up to eight (slopes \SlopeSummary), while kernel estimators slow toward
$n^{-4/(4+d)}$. Its constant is about $2^{-d}$ of that of an unbiased sinc estimator, because BLML estimates $\sqrt f$,
whose band is half that of $f$. We stop at eight dimensions because the constant grows geometrically in $d$ (about $1.85^d$ on
our test densities), so the $n^{-1}$ regime starts at ever larger $n$: at $d=16$ a 1\% relative error would already
need about $2\times10^6$ points, and measuring a rate needs a range of sizes beyond that.

\emph{The cut-off, learned while the data stream in} (Sections~\ref{sec:stream}--\ref{sec:grad}). Each batch updates
only the coefficients of the bins it touches, with the normaliser tracked exactly, and the cut-offs follow Adam steps
on the likelihood of each new batch before it is used. The gradient is analytic; smoothing the log-density where $g$ crosses zero and pooling
gradients over a few batches make it usable. An epoch costs about one solver sweep, so learning the cut-off costs little: \CpuSpeedupRange\
times faster than validation search on a laptop at a held-out likelihood within $0.005$ nats per point.

\emph{Where it stands} (Sections~\ref{sec:exp-nonsmooth}--\ref{sec:uci}). On band-limited truths BLML's error is
2 to 18 times lower than a neural spline flow's from $10^4$ points on. Real data are not band-limited, and there BLML
is the strongest nonparametric estimator we tested. On day-ahead electricity prices it leads conditional kernel and
nearest-neighbour estimators by $0.27$--$0.47$ nats per hour and the flow by $0.05$--$0.12$, and its learned cut-off
drops an uninformative covariate on its own. On the standard POWER and GAS benchmarks, fitted to the full training
sets, it leads a tuned KDE by $0.39$ and $1.58$ nats per point and is ahead of Real NVP on both and of MAF on GAS; the
best published flow stays about $0.6$ nats ahead. It loses where its assumption fails sharply: on a density with a
kink, an adaptive kernel estimator and the flow are better. And a kernel estimator given the same streaming bandwidth
learning keeps its slower rate (Section~\ref{sec:kde}), so the gain does not come from tuning alone.

\emph{Scope.} Our setting is dense data: densities that fill their space, where the curse of dimensionality is real
and every estimator that assumes only smoothness must pay for each added dimension. Data that lie near a
low-dimensional manifold in a high-dimensional space, as images and text do, are a different problem, where estimators do well by
learning that structure, and we do not address it. The dimensions we study, up to eight, are where dense data are
common and the curse is already costly. One of our benchmarks, GAS, lies close to a slow trajectory; we report it
because the flow literature does, and BLML follows it with high cut-offs, but our claims do not rest on it.

Our claims are made within the nonparametric class, estimators whose complexity grows with $n$ and whose single
smoothing parameter is chosen from the data: kernel estimators with rule-of-thumb, cross-validated and oracle
bandwidths, higher-order, sinc and adaptive kernels, and nearest neighbours
\citep{silverman1986,scott2015,rudemo1982,bowman1984,abramson1982,loftsgaarden1965}. Models whose architecture is fixed
in advance, a neural spline flow \citep{durkan2019neural}, on prices a Gaussian mixture, gradient boosting, a
mixture density network and TabPFN, and on the benchmarks the published autoregressive and coupling flows, serve as a
reference for what a well-chosen architecture achieves.
\section{Related work}
\label{sec:related}

Our two contributions touch four literatures: estimators that update as data arrive, incremental fitting of
models with latent coefficients, data-driven choice of smoothing, and binning for fast kernel sums. We place the
work in each.

\paragraph{Recursive and online density estimation.} Recursive kernel estimators add each new point as a
kernel with a bandwidth set by a schedule in $n$ \citep{wolverton1969,wegman1979recursive}; online kernel
density estimation keeps a compressed mixture and revises its bandwidth from the current model
\citep{kristan2011online}. Both update a smoothing parameter by a rule rather than by the likelihood of new
data. Here the estimator is the solution of a convex problem, which streaming must track, and the cut-off is
learned from the prequential score.

\paragraph{Incremental fitting.} Incremental EM updates the sufficient statistics of one block of data at a
time and keeps the bound monotone \citep{neal1998em}; online EM replaces them by stochastic approximation
\citep{cappe2009online}. Touched-bin updates are the analogue for the BLML objective: the counts are the
sufficient statistics, a batch updates the coefficients of its own bins, and the normaliser is tracked
exactly. At fixed parameters an epoch is a pass of block coordinate descent, whose convergence on convex
objectives with a separable non-smooth part is classical \citep{tseng2001,nesterov2012cd,wright2015cd}.

\paragraph{Gradient-based selection of smoothing parameters.} The usual data-driven bandwidths (rules of
thumb, least-squares and likelihood cross-validation, adaptive bandwidths;
\citealp{scott2015,rudemo1982,bowman1984,abramson1982}) refit the estimator for each candidate.
Hyper-gradient methods differentiate a validation loss through the training procedure
\citep{maclaurin2015hyper} or through its optimality conditions \citep{pedregosa2016hyper}. Our
one-sweep response is the cheapest such approximation: the derivative of one coordinate update, taken on the
bins a batch touches.

\paragraph{Binning.} Binning data on a lattice to compute kernel estimators fast is standard
\citep{silverman1982fft,wand1994fast}; BLML's lattice is set far below the Nyquist spacing
(Section~\ref{sec:binned}; \citealp{agarwal2017blml}), and its sinc kernel, with unbounded support, makes the Gram
matrix dense, which is what makes refitting expensive.

\section{Background: band-limited maximum likelihood}
\label{sec:background}

We recall the estimator and its estimating equations, and what the cut-off means, before showing why the
equations become hard to solve beyond two dimensions.

Let $x_1,\dots,x_n\in\mathbb{R}^d$ be i.i.d.\ draws from an unknown density $f$.
BLML \citep{agarwal2017blml} assumes $f=g^2$ with $g$ in the Paley--Wiener space
$\mathcal{V}$ of square-integrable functions whose Fourier transform is supported in
the box $\Omega=\prod_{k=1}^d[-f_{c,k}/2,\,f_{c,k}/2]$, so that $f$ itself is
band-limited to $\prod_k[-f_{c,k},f_{c,k}]$. $\mathcal{V}$ is a reproducing kernel Hilbert
space with the product sinc kernel
\begin{equation}
  K(x,y)=\prod_{k=1}^d f_{c,k}\,\mathrm{sinc}\big(f_{c,k}(x_k-y_k)\big),
  \qquad \mathrm{sinc}(t)=\frac{\sin \pi t}{\pi t},
  \label{eq:sinc}
\end{equation}
which satisfies $\int K(x,z)K(z,y)\,dz=K(x,y)$ \citep{yao1967}. Maximizing
$\prod_i g(x_i)^2$ subject to $\|g\|_2=1$ gives
$\hat f(x)=\big(\tfrac1n\sum_i c_i K(x,x_i)\big)^2$, where $c$ solves
$\rho_i(c)=\tfrac1n\sum_j c_j K(x_i,x_j)-1/c_i=0$. The system has $2^n$ solutions,
one per orthant; when $f>0$ the maximizer lies asymptotically in the positive
orthant, which the BLMLTrivial algorithm solves directly, and BLMLQuick speeds this up
by first binning the data on a lattice finer than the Nyquist spacing
\citep{agarwal2017blml}.

In the language of kernel smoothing, the cut-off plays the role of an inverse bandwidth: the kernel $K$ has width
about $1/f_{c,k}$ in coordinate $k$, so raising $f_c$ sharpens the estimate as shrinking $h$ does, and selecting
$f_c$ is bandwidth selection. The model class is narrower than the Sobolev or H\"older classes of classical theory:
a band-limited $\sqrt f$ extends to an entire function, a far smoother class, which is why the rate can be parametric
rather than $n^{-2s/(2s+d)}$ for smoothness $s$. Densities outside the class, smooth or not, are
approximated with a bias set by how fast their spectrum decays (Sections~\ref{sec:exp-normal}
and~\ref{sec:exp-nonsmooth}).

\section{A solver with linear memory}
\label{sec:method}

From three dimensions on, binning no longer shrinks the problem, and a solver that works with the $B\times B$ kernel
matrix needs memory growing as $n^2$. This section removes the matrix: the binned equations are the optimality
condition of a convex objective, which coordinate descent minimises using one-dimensional kernel tables and a cached
product.

\subsection{The binned problem is a convex program}

Quantize each coordinate to a lattice of spacing $h_k$, giving $B\le n$ distinct
bin centers $z_1,\dots,z_B$ with counts $w_b$ ($\sum_b w_b=n$). Points sharing a bin
share a coefficient by symmetry; writing $e_b$ for the total coefficient of bin $b$
and $S_{ab}=K(z_a,z_b)$, the BLML equations become
\begin{equation}
  (Se)_b\,e_b = n\,w_b,\qquad b=1,\dots,B,
  \qquad
  \hat f(x)=\Big(\frac1n\sum_{b=1}^B e_b K(x,z_b)\Big)^2 .
  \label{eq:binned}
\end{equation}

\begin{proposition}\label{prop:convex}
Let $\Phi(e)=\tfrac12 e^\top S e-n\sum_b w_b\log e_b$ on $e>0$.
(i) $\Phi$ is strictly convex and \eqref{eq:binned} is exactly $\nabla\Phi(e)=0$, so the
positive-orthant BLML solution is the unique minimizer of $\Phi$.
(ii) At any $e>0$ with residuals $\rho_b=(Se)_b e_b/(n w_b)-1$,
$\big|\int\hat f-1\big|\le\max_b|\rho_b|$; in particular $\hat f$ integrates to one at
the solution.
\end{proposition}
\begin{proof}
(i) $S$ is a Gram matrix of a positive definite kernel at distinct points, so
$e^\top Se$ is convex; $-\log$ is strictly convex; $\partial\Phi/\partial e_b=(Se)_b-nw_b/e_b$.
(ii) By the reproducing property, $\int\hat f=n^{-2}e^\top Se
=n^{-2}\sum_b e_b(Se)_b=n^{-2}\sum_b nw_b(1+\rho_b)=1+n^{-1}\sum_b w_b\rho_b$.
\end{proof}

The tolerance on $\max_b|\rho_b|$ used to stop the solver is therefore also a bound on
the normalization error of the returned density.

\subsection{Cached kernel tables}

Because bin centers lie on a lattice, $z_{bk}=h_k\,i_{bk}$ with integer $i_{bk}$, every
Gram entry is a product of one-dimensional table lookups,
\begin{equation}
  S_{ab}=\prod_{k=1}^d t_k[\,i_{ak}-i_{bk}\,],\qquad
  t_k[j]=f_{c,k}\,\mathrm{sinc}(f_{c,k}h_k j),\quad |j|\le J_k,
  \label{eq:tables}
\end{equation}
with $J_k$ the lattice extent of dimension $k$. The tables are exact (no
approximation beyond the binning itself) and occupy $\sum_k(2J_k+1)$ numbers. Each
entry costs $d$ lookups and multiplications instead of $d$ transcendental evaluations.

\paragraph{Binning compresses only in low dimension.}
Following \citet{agarwal2017blml} we use $h_k=0.1\,n^{-1/4}/f_{c,k}$, which keeps the
discretization error of order $1/n$. In one dimension this collapses $n$ points into
$B\ll n$ bins, but in $d\ge3$ almost every point occupies its own bin ($B/n\to1$;
Fig.~\ref{fig:complexity}a). In high dimension the role of the lattice is therefore not
compression but exactness of the table lookups \eqref{eq:tables}; the algorithm below
is what makes large $B$ affordable.

\subsection{Block coordinate descent with a cached sum}
\label{sec:binned}

The exact minimizer of $\Phi$ in coordinate $b$ with the others fixed is the positive
root of $S_{bb}e_b^2+r_be_b-nw_b=0$, $r_b=(Se)_b-S_{bb}e_b$, computed stably as
\begin{equation}
  e_b^\star=\begin{cases}
    2nw_b\big/\big(r_b+\sqrt{r_b^2+4S_{bb}nw_b}\big), & r_b\ge 0,\\[2pt]
    \big(-r_b+\sqrt{r_b^2+4S_{bb}nw_b}\big)\big/(2S_{bb}), & r_b<0.
  \end{cases}
  \label{eq:root}
\end{equation}
We keep the vector $u=Se$ as a cached sum. Each sweep visits the bins in random blocks
$I$ of size $B_s$: it forms the $B\times B_s$ column block $S_{:,I}$ from the tables,
computes $e_I^\star$ for the whole block in parallel from the current $u$, takes a step
$e_I\leftarrow e_I+t\,d_I$ with $d_I=e_I^\star-e_I$, and updates $u\leftarrow u+S_{:,I}\,t\,d_I$.
The full Gram matrix is never formed.

\paragraph{Descent safeguard.}
Updating a block in parallel (Jacobi) can overshoot when bins are strongly coupled. In
one dimension each bin interacts with roughly $1/(f_ch)$ neighbors, and without a
safeguard the iteration oscillated (353 sweeps and a non-monotone objective at
$n=3000$). Since each coordinate moves toward its own one-dimensional minimizer,
$\nabla_I\Phi^\top d_I\le0$, so $d_I$ is a descent direction. The change in $\Phi$ along it
is available exactly from quantities already in memory,
\begin{equation}
  \Phi(e+t d)-\Phi(e)=t\,d_I^\top u_I+\tfrac{t^2}{2}\,d_I^\top S_{II}d_I
  -n\sum_{b\in I}w_b\log\!\Big(1+t\frac{d_b}{e_b}\Big),
  \label{eq:dphi}
\end{equation}
and we take the largest $t\in\{1,\tfrac12,\tfrac14,\dots\}$ for which it is non-positive.
All candidate steps are evaluated in one vectorized operation on the GPU, so the
safeguard adds no synchronization. With it the objective decreases monotonically in
every run we report, and the one-dimensional case needs 11 sweeps instead of 353; in
$d\ge2$ the full step is almost always accepted. To avoid drift in the cached sum under
single precision, $u$ is recomputed exactly (float64 accumulation) before convergence
is declared. The solver stops when $\max_b|\rho_b|<\text{tol}$ (default $10^{-4}$) and has
no iteration cap.

\subsection{Cost}

A sweep touches every column block once, so it costs $O(B^2d)$ table lookups and
multiply--adds. Memory is $O(Bd)$ for the integer lattice coordinates, $O(B)$ for
$e,u,w$, $O(BB_s)$ for one Gram block and $O(\sum_kJ_k)$ for the tables: linear in $B$
and hence in $n$, against $4B^2$ bytes for a dense float32 Gram (40\,GB at $B=10^5$).
The number of sweeps $S$ is small and does not grow with $d$ in our experiments
(Section~\ref{sec:experiments}), so total work is $O(S\,B^2d)$.

\subsection{Periodic dimensions}

For a covariate on a circle of period $P$ (hour of day, day of year) we replace the
sinc factor by the normalized Dirichlet kernel
\begin{equation}
  K_P(t)=\frac1P\,\frac{\sin\!\big(\pi(2N+1)t/P\big)}{\sin(\pi t/P)},\qquad
  N=\big\lfloor f_cP/2\big\rfloor,
  \label{eq:dirichlet}
\end{equation}
the reproducing kernel of trigonometric polynomials of degree $N$ on $[0,P)$. The choice
of $N$ keeps the same band convention as \eqref{eq:sinc} ($|\nu|\le f_c/2$ for $g$), and the
factor $1/P$ makes $K_P$ reproducing over one period, so Proposition~\ref{prop:convex}
holds unchanged. $N$ must be an integer: with a fractional mode count the function in
\eqref{eq:dirichlet} is neither periodic nor positive semidefinite (at $f_c=0.134$,
$P=24$ its Gram matrix has a negative eigenvalue of 5\% of the largest), and $\Phi$ is no
longer convex. On the lattice the periodic Gram factor is circulant, and its table is
indexed by the minimum circular distance.

\subsection{GPU implementation}

The only custom GPU code is one CUDA kernel that writes the $B\times B_s$ block
$S_{:,I}$: one thread per entry, a loop over dimensions with a table lookup per
dimension (\eqref{eq:tables}, or the circular index for periodic dimensions), compiled
once per process with NVRTC through CuPy \citep{okuta2017cupy}. The block update,
safeguard and cached-sum update are a handful of CuPy array operations. Prediction
evaluates the continuous kernel (not the lattice) at query points in blocks. The CPU
reference implementation uses the same tables with Numba \citep{lam2015numba}. Both
precisions are supported; on an A100, float32 and float64 run at comparable
speed (Section~\ref{sec:experiments}) because the solver is limited by memory traffic and
kernel launches rather than arithmetic.
\section{Learning the cut-off while streaming}
\label{sec:learn}

The solver fits BLML at a given cut-off. Choosing the cut-off is the expensive part: the standard way refits the
estimator tens of times. This section replaces those refits by a single streaming fit that learns the cut-off on the
way, in four steps: the baseline search it replaces, coefficient updates that touch only a batch's bins, gradients of
the held-out likelihood in the cut-off, and the resulting cost.

\subsection{The baseline: one-shot selection}
\label{sec:oneshot}

The baseline chooses the cut-off vector $f_c$, one frequency per dimension, by maximizing the
log-likelihood of held-out validation points $V$ under a fit to the remaining training points,
\begin{equation}
  \mathrm{VLL}(f_c)=\frac{1}{|V|}\sum_{i\in V}\log\hat f_{f_c}(x_i),
  \label{eq:vll}
\end{equation}
or the conditional log-likelihood for conditional models. The search runs in coordinates scaled by each
dimension's interquartile range, so that one starting cut-off ($0.2$ per scaled unit) suits every
dimension; the result is mapped back to the original units.

The search has two stages, a shared search over a grid of cut-off volumes and a per-dimension refinement, with
periodic dimensions moving in whole harmonics (Appendix~\ref{app:oneshot}).

\paragraph{Cost.} Every candidate is a fit to tolerance on all training points, and a selection takes
15--45 of them: with $F$ candidates of $s$ sweeps, $O(F\,s\,d\,n^2)$ in the worst case of one point per
bin, plus the validation scores. The per-column refinement makes $F$ grow linearly in $d$, so the cost is
quadratic in the dimension, $O(s\,d^2n^2)$. Streaming (Sections~\ref{sec:stream}--\ref{sec:grad}) replaces
this search after a burn-in on part of the data, where the same search runs once.
\subsection{Streaming at a fixed cut-off}
\label{sec:stream}

A streaming fit is useful only if each batch costs far less than a refit, yet the result is still the exact estimate.
We first show how to achieve both at a fixed cut-off; learning the cut-off on the way is Section~\ref{sec:grad}.

\paragraph{Why not refit every bin.} The obvious streaming update refits all held bins at each batch, even
for one sweep. With $n_t$ bins held it costs $O(d\,n_t^2)$ per batch and $O(d\,n^3/b)$ per epoch of
batches of $b$ points: cubic in $n$, and worse for small batches. It is competitive only when there are few
batches.

\paragraph{Touched-bin updates.} After burn-in (an exact fit on the first $n_0$ points), a batch of $b$
points updates only the coefficients of the bins it falls in, $I$ with $|I|\le b$; the rest of the lattice
is not visited. The batch's counts are added and the rows $(Se)_I$ are computed fresh against all $B$ bins; nothing of
size $B$ is carried, so nothing goes stale when a kernel parameter moves. Each $e_b$, $b\in I$, then takes one
coordinate step of the solver (Section~\ref{sec:binned}), in blocks with its line search. New bins start at
zero and form blocks of their own: from zero the step goes to the positive root whatever the sign of $R_b$, and
any positive value lowers $\Phi$.

\paragraph{The normaliser.} $Q=e^\top Se$ is tracked exactly through the updates,
$Q\leftarrow Q+2\,\delta^\top(Se)_J+\delta^\top S_{JJ}\delta$ for a block $J$ with step $\delta$, and the
coefficients are rescaled to $e^\top Se=n^2$, which keeps $\hat f$ a density at every batch.

\paragraph{Epochs.} A second pass revisits the same data in a new random order, continuing from the
current coefficients; counts are not added again, and every bin is met by every batch's rows.

\paragraph{Stopping rule.} With a tolerance, the number of epochs is a cap and the pass stops when the
solver's own test, $\max_b|\rho_b|<\epsilon$ (Proposition~\ref{prop:convex}), is met. The residual is checked either exactly, one extra pass of $Se$
over all bins, or as a running maximum over the epoch of each touched bin's residual before its update,
which is free because the rows $(Se)_I$ are computed anyway; the running check counts from the second
epoch, when every bin has been seen. The stopping rule needs every bin to take its step, so it is not
combined with partial block updates or with frozen bins.

\paragraph{Lattice.} The lattice for raw points follows $h\propto n^{-1/4}/f_c$, but is rebuilt only when
the target width moves by more than a factor $2^{1/4}$ (the data double, or $f_c$ moves by 19\%), so
$O(\log n)$ times in all; all runs in Section~\ref{sec:exp} use this tolerance. A factor of 2 works as well
(Appendix~\ref{app:rebin}). A rebuild starts each new bin at the current fit's stationary value $w_b/g(z_b)$
where $g(z_b)>0$, and at zero elsewhere, and runs two full sweeps. (A negative start lies outside the
domain $e_b>0$, where the line search sees the cost of a step toward the positive root but not the
barrier it crosses, and rejects it.) A periodic column with a known period is folded (points wrapped to
$[0,P)$, $P/h$ an integer); with a trainable period it is binned on the raw coordinate and its table row is
the Dirichlet kernel at raw offsets, which is $P$-periodic by itself, so a period step changes only that
row.

\paragraph{Convergence at fixed parameters.} After the first epoch the counts are complete and, with the
parameters fixed, the lattice no longer changes, so each later epoch minimises the same objective
$\Phi$ of Proposition~\ref{prop:convex}.

\begin{proposition}
\label{prop:conv}
Fix the kernel parameters and the lattice, with $S$ positive definite on the held bins (true for a sinc
kernel at distinct lattice points, whose spectrum is positive on an interval), and let every held bin (each
with $w_b>0$) be updated at least once per epoch by an exact coordinate step, followed by the rescaling to $e^\top Se=n^2$. Then $\Phi$ decreases
at every step and the coefficients converge to the unique minimiser of $\Phi$, the exact fit.
\end{proposition}

\begin{proof}[Sketch]
$\Phi$ is strictly convex on $e>0$ and its sublevel sets are compact (the quadratic term dominates as
$\|e\|\to\infty$, the log term as any $e_b\to0$), so the minimiser $e^*$ is unique. The coordinate step
minimises $\Phi$ exactly along $e_b$ (it is the positive root of $\partial\Phi/\partial e_b=0$), and the
rescaling minimises it exactly along $e$, since $\Phi(ce)=\tfrac{c^2}2e^\top Se-n^2\log c+\text{const}$ is
least at $c^2e^\top Se=n^2$. $\Phi$ is a smooth function plus a separable term, so block coordinate descent
with exact minimisation in an essentially cyclic order (every coordinate within each epoch) converges to
$e^*$ \citep{tseng2001}.
\end{proof}

The implementation updates blocks of coordinates together with a line search on $\Phi$, which keeps the
decrease monotone but is not covered by the proposition as stated; Section~\ref{sec:fixed} checks that it
reaches the exact fit, to the solver's tolerance.
\subsection{Learning the cut-off}
\label{sec:grad}

With the coefficients kept exact batch by batch, each new batch offers a free out-of-sample test of the current
cut-off before it is used. We learn the kernel parameters from that prequential score: below we describe what is
differentiated (the kernel classes), the score and its gradient, the corrections that make the gradient usable, and
how a step is applied.

\paragraph{Kernel classes.} The product kernel is a product of units: a sinc column, a periodic (Dirichlet)
column, or a radial group of columns sharing one band. Each unit has the form
$K(t)=\prod_kf_k\,p(\|f\odot t\|)$, where $p$ is the band set's profile at unit cut-off. Scaling the unit's
cut-off by $e^{\theta}$ gives
\begin{equation}
  \frac{\partial K}{\partial\theta}=\prod_kf_k\,\big(D\,p(s)+s\,p'(s)\big),\qquad s=\|f\odot t\|,
\end{equation}
with $D$ the unit's dimension: $f\cos(\pi ft)$ for sinc, and for a ball
\[s\,p'(s)=-\pi c\,s^{1-D/2}J_{D/2+1}(\pi s).\] A Dirichlet column of period $P$ with $M=2N+1$
harmonics has $K(t)=\frac1PD_M(2\pi t/P)$; its cut-off moves in whole harmonics (no gradient), and at
fixed $N$ the period has $\partial K/\partial\log P=-\frac1P\big(D_M(u)+uD_M'(u)\big)$, $u=2\pi t/P$. Each
class provides $K$ and $\partial K/\partial\theta$ both at arbitrary points and as lattice table rows, and
declares its parameters with a type (scale, integer) and default trainability; a fixed parameter is used as
given, a trainable one with a value starts there.

\paragraph{Prequential score.} Each batch is scored by the current model before it is used,
\begin{equation}
  \ell_t=\frac1m\sum_{x\in\text{batch }t}\log\hat f_{t-1}(x),
\end{equation}
or with $\log\hat f(x_{\mathrm{dep}}\mid x_{\mathrm{rest}})$ for conditional models (joint over marginal),
the prequential score \citep{dawid1984prequential}. It is an out-of-sample estimate of the current model's
log-likelihood, and its gradient in the kernel parameters drives their updates. A revisited batch (a later
epoch) is already in the fit, so to keep its score and gradient out of sample it is scored leave-batch-out:
its share $c_b/w_b$ of each touched coefficient is removed before scoring, $c_b$ its count in bin $b$, which
is exact to first order because $e_b$ is close to proportional to the bin's count.

\paragraph{Gradient of the prequential score.} With $e$ fixed and $Z=e^\top Se/n^2$,
\begin{equation}
  \frac{\partial}{\partial\theta}\Big[\log g(x)^2-\log Z\Big]
  =\frac{2\,\partial_\theta g(x)}{g(x)}-\frac{e^\top(\partial_\theta S)e}{n^2Z}.
\end{equation}
$\partial_\theta g$ needs the kernel derivative at the batch points, for all units at once: with
$K=\prod_uK_u$, prefix and suffix products over the units give $K$ and every $\partial K/\partial\theta_u$
in $O(d)$ per point--bin pair. The normaliser term $e^\top(\partial_\theta S)e=\sum_be_b(\partial_\theta Se)_b$
is a sum over all bins; we estimate it on the bins of a sample of held points (bin $b$ with
probability $w_b/n$; the same sample estimates the change in $Q$ below), as the ratio of the weighted sums of $e_b(\partial_\theta Se)_b$ and $e_b(Se)_b$, so a
step costs $O(d\,b\,B)$, not a lattice pass. The batch's own bins would not do: when most points have bins of
their own, the bins of a new batch are new, with $e_b=0$.

\paragraph{The fixed-coefficient gradient is biased.} The envelope theorem makes the fixed-$e$ derivative
exact for the training objective, but the prequential score is not that objective. The coefficients carry
sampling noise that the band hides; raising $f_c$ with $e$ fixed exposes it, while a refit would absorb it.
The fixed-$e$ gradient therefore understates the gain from a higher cut-off: in our first streaming
design, which refitted every bin at each batch, it stopped one Dirichlet harmonic short and drove a sinc cut-off to
half its selected value. With touched-bin updates the bias matters only when a step follows every batch without
smoothing; with smoothing and gradients pooled over three batches, the default, correcting it changes the held-out
score by $-0.0015$ nats per point on average over five seeds, within the spread between seeds
(Appendix~\ref{sec:ablation}), so it is off by default. The correction differentiates through one sweep. Each coefficient update is the positive root of $S_{bb}t^2+R_bt-nw_b=0$,
$R_b=(Se)_b-S_{bb}e_b$, so at $t=e_b$
\begin{equation}
  \frac{\partial t}{\partial\theta}=-\frac{e_b\,\partial R_b+e_b^2\,\partial S_{bb}}{2S_{bb}e_b+R_b},
\end{equation}
with $\partial R_b=(\partial S\,e)_b-\partial S_{bb}\,e_b$, taken on the touched bins (the only
coefficients a batch moves), from the rows $(\partial S\,e)_I$ computed with $(Se)_I$.

\paragraph{Smoothing.} A real $g$ crosses zero in the tails, and $\log g^2$ is singular there: the few
batch points in such lobes ($\hat f\sim10^{-10}$) carry per-point gradients $\sim10^4$ and set the
direction of every step. We take the gradient of $\log(\hat f+\varepsilon)$, $\varepsilon=10^{-3}$ times the
median density at the bins (where $g=(Se)_b/n$, free from the fit), which bounds those points without
moving the others.

\paragraph{Moving a parameter.} A parameter step changes $Q$ everywhere. We estimate the ratio
$Q(\theta')/Q(\theta)$ on the same sample of held bins, and correct the tracked $Q$ at every batch toward the
same estimator's value of $Q$ (a running average with weight $0.1$), so estimation errors do not accumulate.
The coefficients are not refitted when a parameter moves; later batches and epochs bring them along.

\paragraph{Updates.} The update rule follows each parameter's declared type, not its meaning: scale
parameters move by Adam \citep{kingma2015adam} on $\log\theta$ (learning rate $0.1$), with the gradient
pooled over $3$ batches, each contributing before it is fitted; integer parameters (Dirichlet harmonics)
move one step at a time once the pooled paired gain of the move exceeds one standard error. A hill-climb on
the scale parameters (steps of $5\%$, the same paired test) is available in place of Adam.

\paragraph{Periods.} A period can be learned only for a conditioning column of a conditional density. In a
joint fit the marginal of $t\bmod P$ is a density on a circle of length $P$ and carries $-\log P$, so the
likelihood always prefers a shorter period (in our runs a joint fit walked from 22\,h to 8\,h once the
harmonics in $t$ dropped to zero); in $p(y\mid t)$ that marginal cancels. The data must also span several
periods: on one period, e.g.\ hour of day, any $P$ beyond the data's range wraps nothing and fits about as
well.
\subsection{Cost}
\label{sec:cost}

How much does the cut-off cost when it is learned on the way? With $n_t$ bins held, a batch of $b$ points costs
$O(d\,b\,n_t)$: the prequential score and its gradient (one pass over the bins; the derivatives in every cut-off cost
no extra factor of $d$, Section~\ref{sec:grad}), the rows $(Se)_I$ and their derivatives, and one block update. In the
first epoch the lattice grows with the stream, so its batches cost about $n^2/2$ pair evaluations in all; a later
epoch costs $n^2$. With a burn-in selection on $n_b$ points,
\[
  \text{streaming: } O\bigl(E\,d\,n^2 + F\,s\,d\,n_b^2\bigr), \qquad
  \text{one-shot selection: } O\bigl(F\,s\,d\,n^2\bigr),
\]
for $E$ epochs and $F$ candidate fits of $s$ sweeps: streaming replaces $F\,s$, one to a few hundred, by about $2E$.
Memory is that of the solver, linear in $n$. The fixed-parameter path, GPU precision and the kernel tables are in
Appendix~\ref{sec:costdetail}.
\section{Why the rate need not depend on dimension}
\label{sec:rate}

Every estimator that assumes only smoothness slows down as $d$ grows; why should BLML not? The following heuristic,
not a proof, gives a reason and two predictions that the experiments test (Appendix~\ref{app:conjecture} states it
as a conjecture and outlines a proof). Suppose $f$
has most of its mass in a box with side lengths $R_k$. Functions band-limited to
$\Omega=\prod_k[-f_{c,k}/2,f_{c,k}/2]$ and essentially supported in that box form,
to good approximation, a linear space of dimension
$p_{\text{eff}}\approx\prod_k f_{c,k}R_k$. This is the multidimensional version of the
time--bandwidth count of \citet{landau1962prolate}: the time- and band-limiting operator has about
that many eigenvalues near one and the rest near zero \citep{landau1980eigenvalue,slepian1976bandwidth}.
BLML's solution is written with $n$ coefficients, one per sample, but once $n\gg p_{\text{eff}}$
the $n$ kernel functions are highly redundant (they span essentially the same
$p_{\text{eff}}$-dimensional space), so the number of parameters that matter statistically is
$p_{\text{eff}}$, not $n$. For a product density a natural width is $R_k=1/\int f_k^2$. Within that space BLML is a maximum
likelihood estimator in $p_{\text{eff}}$ effective parameters, so standard parametric
asymptotics suggest an integrated error of order $p_{\text{eff}}/n$. The exponent of $n$
is then $-1$ for every $d$, and dimension enters only through the constant
$p_{\text{eff}}$, which grows geometrically in $d$. Two consequences are testable: the
log--log slope of the error against $n$ should stay near $-1$ as $d$ grows, and the
error at fixed $n$ should grow with $d$. A KDE, by contrast, has bias of order $h^2$
and variance of order $(nh^d)^{-1}$, giving slope $-4/(4+d)$. The heuristic also
indicates where it fails: when $n\lesssim p_{\text{eff}}$ the estimator is not yet in its
asymptotic regime, so at high $d$ the $-1$ slope should appear only at large $n$.

\paragraph{Why we stop at eight dimensions.} The rate is free of dimension; the constant is not, and that is BLML's
form of the curse of dimensionality. With the measured constant of about $1.85$ effective parameters per dimension on
T1 (Section~\ref{sec:exp-rate}), the error is roughly $1.85^d/n$, so a relative ISE of 1\% needs about $1.4\times10^4$
points at $d=8$, $1.6\times10^5$ at $d=12$, $1.9\times10^6$ at $d=16$, $2.6\times10^8$ at $d=24$ and $3.5\times10^{10}$ at
$d=32$. Measuring a rate needs more: a grid of sample sizes reaching well past that point, with replicates. The solver
costs $O(n^2d)$ and fits $10^6$ points at $d=8$ in about ten minutes on one A100, so a single fit at the $1\%$ point
takes about an hour at $d=16$ and years at $d=24$. Eight is the largest dimension at which the $n^{-1}$ regime can be
reached and traced across a grid of sample sizes with replicates; above it, any estimator that assumes only smoothness
is pre-asymptotic at every sample size one could collect, and what decides its accuracy is structure the smoothness
class does not encode (Section~\ref{sec:discussion}).

\section{Experiments: the rate in high dimension}
\label{sec:experiments}

Does the prediction hold? We measure BLML's error against $n$ and $d$ on truths inside and outside its class.

\subsection{Setup}

\begin{figure}[t]
  \centering
  \includegraphics[width=\textwidth]{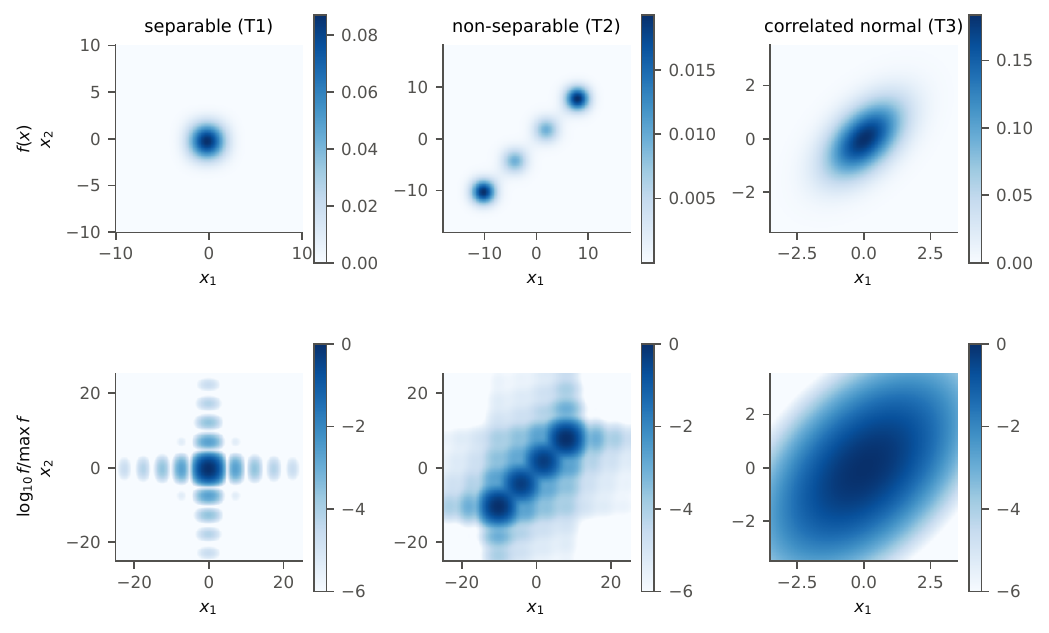}
  \caption{The truths in two dimensions, on a linear scale (top) and as $\log_{10}f/\max f$ (bottom). T1: a product
  of $\mathrm{sinc}^4$ bumps; T2: four bumps along the diagonal whose ripples do not factor; T3: a correlated normal,
  not band-limited.}
  \label{fig:truths}
\end{figure}

\paragraph{Truths.} \textbf{T1} is the product $f(x)=\prod_kf_1(x_k)$ of the base paper's
$f_1(t)=0.15\,[\mathrm{sinc}^4(0.2t)+\mathrm{sinc}^4(0.2t+0.1)]$, band-limited with $f_c^{\text{true}}=0.4$ per dimension.
\textbf{T2} is non-separable, $f=h^2/\|h\|^2$ with $h$ a positive sum of four products of
$\mathrm{sinc}^2(t)+\mathrm{sinc}^2(t+0.5)$ along the diagonal, so $\sqrt f$ is band-limited with the same cut-off.
Both are strictly positive with $|x|^{-4}$ tails, as BLML's consistency theory requires. \textbf{T3} is a correlated
normal (unit variances, correlation $0.5$), smooth but not band-limited (Fig.~\ref{fig:truths}). All are sampled
exactly (Appendix~\ref{app:setup}).

\paragraph{Error.} The main metric is the relative integrated squared error (ISE) $\int(\hat f-f)^2/\int f^2$, by
quadrature for $d\le2$ and by importance sampling from a defensive Cauchy mixture above; we also report the
Kullback--Leibler divergence on draws from $f$ (Appendix~\ref{app:setup}).

\paragraph{Estimator classes.} We distinguish three classes and make our claims within the first. A
\emph{nonparametric} estimator assumes no functional form and lets its complexity grow with $n$, with one smoothing
parameter chosen from the data: BLML (one coefficient per bin, the cut-off its smoothing parameter), the kernel
estimators and $k$-nearest neighbours. A \emph{flexible parametric} model fixes its architecture before seeing the
data: the Gaussian mixture, the mixture density network, the neural spline flow and the pretrained TabPFN. A
\emph{semiparametric} model combines a parametric output family with a nonparametric part, as XGBoostLSS does with a
Gaussian whose mean and variance are boosted trees. The last two classes are a reference, not like-for-like
competitors.

\paragraph{Estimators.} BLML at $f_c^{\text{true}}$ and $2f_c^{\text{true}}$ (float64, tolerance $10^{-4}$); the
Gaussian product KDE with Scott's rule, the least-squares cross-validated bandwidth \citep{rudemo1982,bowman1984} and an
\emph{oracle} bandwidth minimising the true error, the best case for a KDE; the sixth-order and sinc KDEs
\citep{hall1988order,davis1975}; Abramson's adaptive KDE \citep{abramson1982}; and $k$-NN with $k=\sqrt n$
\citep{loftsgaarden1965}. The sixth-order and sinc KDEs go negative (the sinc KDE at up to half the sample points in
eight dimensions) and $k$-NN does not integrate to one, so these \emph{signed} estimators appear as references only.

\paragraph{Design.} $d\in\{1,2,3,4,6,8\}$, $n\in\{10^3,3\cdot10^3,10^4,3\cdot10^4,10^5\}$, 8 replicates per cell (4
at $n=10^5$ for $d\ge4$), and BLML alone at $n=3\times10^5$ and $10^6$ for T1 at $d=4,6,8$. Slopes are least-squares
fits of log mean error on $\log n$, with bootstrap intervals over replicates. All runs use one A100.

\subsection{Scaling to high dimension}
\label{sec:exp-scale}

Without the solver none of what follows could be run: a dense float32 Gram matrix exceeds the A100's 40\,GB beyond
$n\approx10^5$ (Fig.~\ref{fig:complexity}b), and from $d=3$ on binning leaves $B=n$ (at $n=3000$, $B/n=\BoverNdOne$ in
one dimension, $0.92$ in two). The solver's peak memory grows linearly with $n$ and hardly with $d$, \MemPeakDEight\
at $n=10^5$ in eight dimensions (Appendix Table~\ref{tab:mem}). Fit time grows roughly linearly in $n$ until the GPU
is saturated near $n=3\times10^4$ ($0.07$\,s at $n=10^3$, $0.9$\,s at $3\times10^4$, $d=3$) and then toward the
quadratic regime ($3.2$--$5.4$\,s at $n=10^5$; the largest fits, $n=10^6$ at $d=8$, take about \FitTimeBig), and
roughly linearly in $d$. Refining the lattice four-fold changes the error by under 1\% (Appendix Table~\ref{tab:bin}).

\subsection{Band-limited truths: the rate and the constant}
\label{sec:exp-rate}

\begin{figure}[t]
  \centering
  \includegraphics[width=\textwidth]{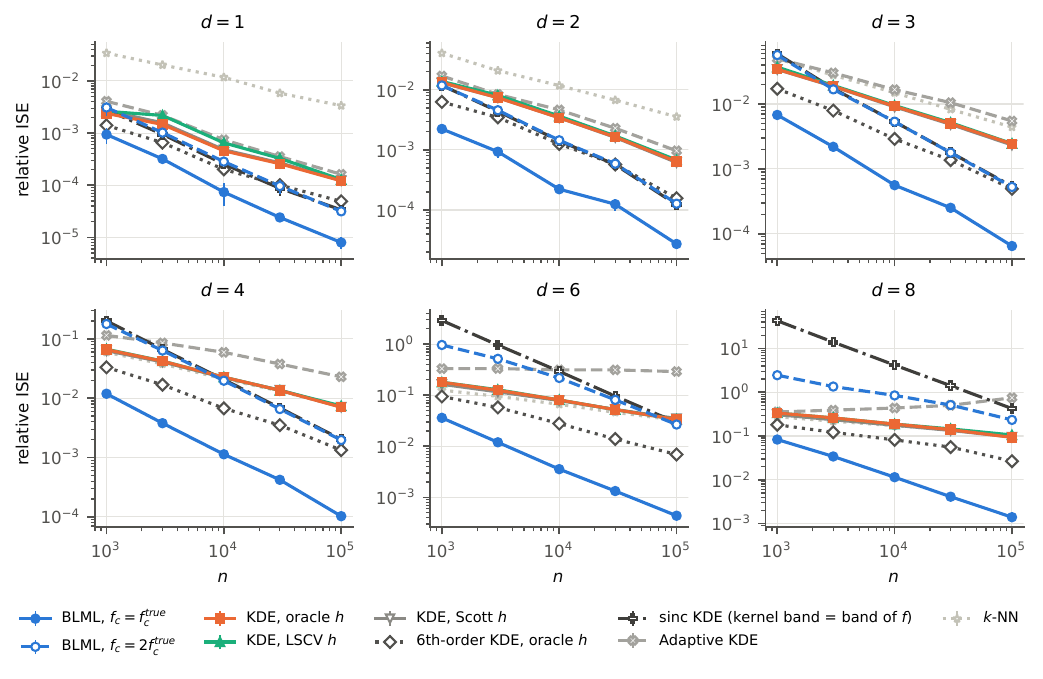}
  \caption{Relative integrated squared error against $n$ for the separable truth T1
  (mean $\pm$ s.e.\ over replicates).}
  \label{fig:rate}
\end{figure}

\begin{figure}[t]
  \centering
  \includegraphics[width=\textwidth]{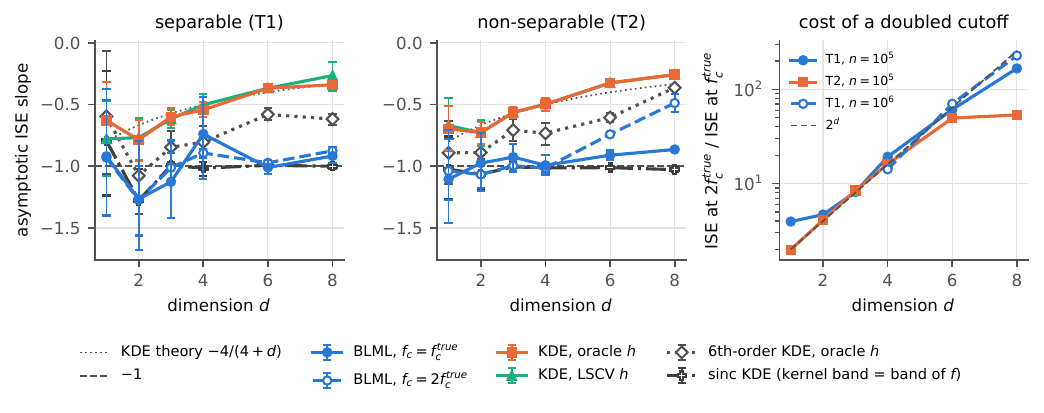}
  \caption{Left, centre: asymptotic slope of log relative ISE against $\log n$, fitted over the
  two largest $n$ ($3\times10^4$ and $10^5$; for BLML on T1 in $d=4,6,8$, the large-$n$ runs at
  $3\times10^5$ and $10^6$), with 95\% bootstrap intervals. Dashed: $-1$; dotted: the KDE rate
  $-4/(4+d)$. Slopes over all $n$ are in Table~\ref{tab:slopes}. Right: the cost of a cutoff twice
  too high, as the ratio of BLML's error at $2f_c^{\text{true}}$ to that at $f_c^{\text{true}}$; dashed:
  $2^d$.}
  \label{fig:slopes}
\end{figure}

\paragraph{On band-limited truths BLML has the lowest error everywhere.} At $f_c=f_c^{\text{true}}$ it beats every
estimator at every $n$ and $d$ on T1 (Fig.~\ref{fig:rate}, Table~\ref{tab:err}). At $n=10^5$ its relative ISE is
$15$--$77\times$ below the best estimator that returns a density (the oracle-bandwidth Gaussian KDE) and $4.2$--$19\times$
below the best signed one. Its slope over all $n$ stays near $-1$ through $d=6$ ($-1.05$, $-0.94$, $-1.00$, $-1.02$,
$-0.96$) and is $-0.89$ at $d=8$ (Table~\ref{tab:slopes}, Fig.~\ref{fig:slopes}), while the Gaussian KDE's flattens
from $-0.67$ to $-0.28$ and the sixth-order KDE's from $-0.75$ to $-0.40$. The non-separable T2 gives the same picture
with smaller margins ($3.3$--$15\times$ at $n=10^5$; slopes $-0.91$ to $-0.72$), except at small $n$ in high
dimension, where the sixth-order KDE, which is not a density, does better.

\begin{figure}[t]
  \centering
  \includegraphics[width=\textwidth]{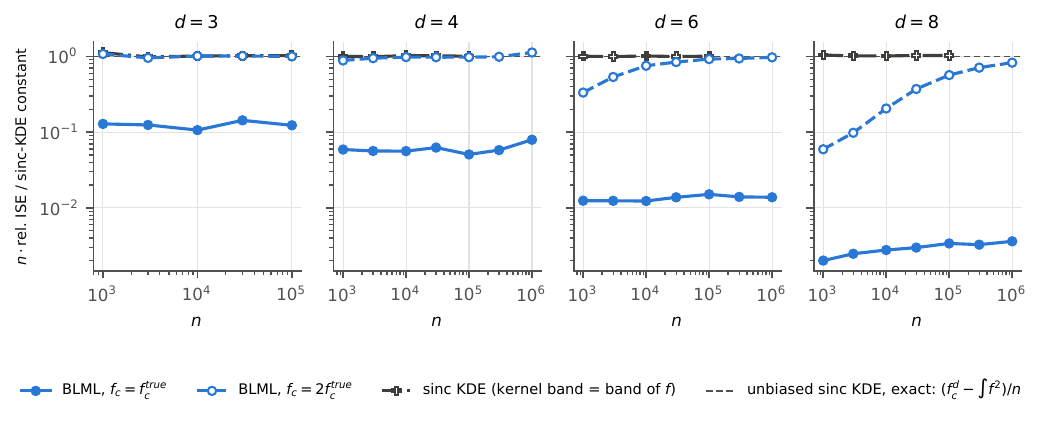}
  \caption{$n\cdot$ relative ISE divided by the exact constant $f_c^d/\int f^2-1$ of the
  unbiased sinc KDE (dashed line at 1), T1. A flat line is an $n^{-1}$ rate; its height is the
  constant. BLML at $f_c^{\text{true}}$ sits near $2^{-d}$; at $2f_c^{\text{true}}$ it approaches the
  sinc KDE's constant from below. BLML points beyond $n=10^5$ are from the large-$n$ runs.}
  \label{fig:scaled}
\end{figure}

\paragraph{The constant, not the rate, is what sets BLML apart.} The sinc KDE whose band equals that of $f$ is
unbiased, with MISE exactly $(f_c^d-\int f^2)/n$, and its slopes are $-1.00$ in every dimension, so $n^{-1}$ alone
does not distinguish BLML. Its constant does (Fig.~\ref{fig:scaled}): BLML sits on flat lines at
$0.11$--$0.13$, $0.05$--$0.08$, $0.012$--$0.015$ and $0.002$--$0.003$ of the sinc KDE's constant for $d=3,4,6,8$, against
$2^{-d}=0.125$, $0.0625$, $0.016$ and $0.004$. This is the effective dimension of Section~\ref{sec:rate}: BLML estimates
$\sqrt f$, whose band is half that of $f$, so it needs about $2^{-d}$ as many effective parameters. The count is
quantitatively right: $f_cR=0.4/0.212=1.89$ per dimension predicted, $2.0$, $1.85$ and $1.85$ measured at $d=4,6,8$.
It also shows the limit of any such guarantee: if the per-dimension factor held, at $d=1024$ the constant would be $1.89^{1024}\approx10^{282}$.

\paragraph{A cut-off that is too high costs constant, not rate.} At $2f_c^{\text{true}}$ BLML's constant approaches the
sinc KDE's from \emph{below} ($0.06\to0.83$ at $d=8$ from $n=10^3$ to $10^6$): before the asymptote, at
$n\lesssim p_{\text{eff}}$, the error is lower than the $n^{-1}$ line, which is why slopes fitted over
$10^3\le n\le10^5$ flatten at high $d$. In error at fixed $n$, a doubled cut-off costs about $2^d$ ($4.0$, $4.7$,
$8.1$, $19$, $61$, $167$ for $d=1,2,3,4,6,8$ at $n=10^5$), and the minimum of the error stays at the true cut-off as $n$
grows (Fig.~\ref{fig:fc-sweep}, left). BLML also has the lowest Kullback--Leibler divergence among valid densities on
T1 ($6$--$72\times$ below the oracle Gaussian KDE at $n=10^5$; Appendix~\ref{app:kl}).

\subsection{A smooth truth that is not band-limited}
\label{sec:exp-normal}

\begin{figure}[t]
  \centering
  \includegraphics[width=\textwidth]{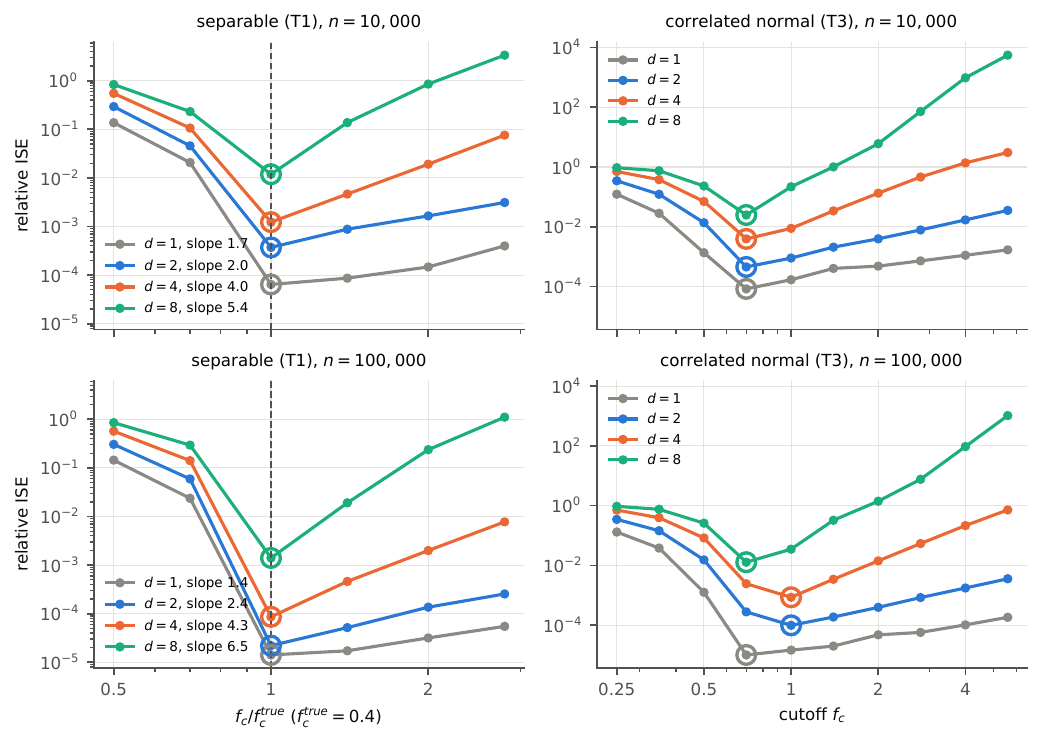}
  \caption{Relative ISE against the cutoff, at $n=10^4$ (top) and $n=10^5$ (bottom); circles mark
  each curve's minimum. Left: T1, with the cutoff relative to the true one (dashed). The minimum
  stays at $f_c^{\text{true}}$ as $n$ grows, and the slope of the right branch (in the legend) grows
  with $d$. Right: T3, which has no true cutoff; the minimum moves to higher cutoffs as $n$ grows.
  Left of each minimum the error is dominated by truncation bias, right of it by variance.}
  \label{fig:fc-sweep}
\end{figure}

What if the band limit is false? For the Gaussian T3 the truncation bias falls fast with $f_c$ ($5\times10^{-5}$ at
$f_c=1$, $7\times10^{-18}$ at $2$) while the variance constant grows as about $f_c^d$, so the cut-off becomes a
bias--variance knob (Fig.~\ref{fig:t3-cutoffs}). A high cut-off ($f_c\ge2$) keeps the $n^{-1}$ rate but pays about
$2^d$ in the constant: at $d=8$ it is $30\times$ worse than the sixth-order KDE. A low cut-off ($f_c=0.7$) wins at
small $n$ in every dimension ($0.024$ against $0.10$ for the sixth-order and $0.22$ for the Gaussian KDE at $d=8$,
$n=10^4$, both with oracle bandwidths) but levels off at its truncation bias. The best cut-off moves up as $n$ grows
(Fig.~\ref{fig:fc-sweep}, right), and chosen per $n$, BLML has the lowest error in every T3 cell: $2.7$--$12\times$
below the best valid estimator. The cut-off must then be learned from data, which Section~\ref{sec:learn} does.

\begin{table}[t]
  \centering\small
  \caption{Relative ISE at the largest $n$ ($10^5$) per dimension; bold marks the best
  estimator. T3 has no true cutoff; its BLML rows use $f_c=2$ and $4$.}
  \label{tab:err}
  \fitwidth{%
\begin{tabular}{lcccccc}
\toprule
Estimator & $d=1$ & $d=2$ & $d=3$ & $d=4$ & $d=6$ & $d=8$ \\
\midrule
\multicolumn{7}{l}{\textit{T1, largest $n$ per $d$}} \\
BLML, $f_c=f_c^{true}$ & \textbf{7.97e-6} & \textbf{2.72e-5} & \textbf{6.50e-5} & \textbf{1.03e-4} & \textbf{4.37e-4} & \textbf{1.40e-3} \\
BLML, $f_c=2f_c^{true}$ & 3.15e-5 & 1.27e-4 & 5.28e-4 & 1.97e-3 & 2.67e-2 & 2.33e-1 \\
KDE, oracle $h$ & 1.21e-4 & 6.39e-4 & 2.43e-3 & 7.07e-3 & 3.35e-2 & 9.30e-2 \\
KDE, LSCV $h$ & 1.28e-4 & 6.79e-4 & 2.46e-3 & 7.43e-3 & 3.35e-2 & 1.06e-1 \\
KDE, Scott $h$ & 1.23e-4 & 6.18e-4 & 2.34e-3 & 7.10e-3 & 3.50e-2 & 9.47e-2 \\
6th-order KDE, oracle $h$$^\dagger$ & 4.90e-5 & 1.57e-4 & 4.93e-4 & 1.33e-3 & 6.90e-3 & 2.66e-2 \\
sinc KDE (kernel band = band of $f$)$^\dagger$ & 3.36e-5 & 1.22e-4 & 5.42e-4 & 2.03e-3 & 2.91e-2 & 4.24e-1 \\
Adaptive KDE & 1.62e-4 & 9.68e-4 & 5.48e-3 & 2.30e-2 & 2.90e-1 & 7.48e-1 \\
$k$-NN$^\ddagger$ & 3.36e-3 & 3.57e-3 & 4.41e-3 & 7.26e-3 & 3.24e-2 & 1.03e-1 \\
\multicolumn{7}{l}{\textit{T2, largest $n$ per $d$}} \\
BLML, $f_c=f_c^{true}$ & \textbf{8.86e-5} & \textbf{1.89e-4} & \textbf{3.97e-4} & \textbf{9.34e-4} & \textbf{4.86e-3} & \textbf{2.02e-2} \\
BLML, $f_c=2f_c^{true}$ & 1.76e-4 & 7.73e-4 & 3.35e-3 & 1.50e-2 & 2.41e-1 & 1.08e+00 \\
KDE, oracle $h$ & 2.93e-4 & 1.60e-3 & 5.25e-3 & 1.42e-2 & 5.84e-2 & 1.50e-1 \\
KDE, LSCV $h$ & 3.27e-4 & 1.64e-3 & 5.25e-3 & 1.42e-2 & 5.84e-2 & 1.50e-1 \\
KDE, Scott $h$ & 2.89e-3 & 6.03e-2 & 2.20e-1 & 4.54e-1 & 8.24e-1 & 9.72e-1 \\
6th-order KDE, oracle $h$$^\dagger$ & 1.59e-4 & 5.23e-4 & 1.51e-3 & 3.59e-3 & 1.68e-2 & 5.77e-2 \\
sinc KDE (kernel band = band of $f$)$^\dagger$ & 1.77e-4 & 7.80e-4 & 3.35e-3 & 1.55e-2 & 3.25e-1 & 6.76e+00 \\
Adaptive KDE & 3.17e-4 & 2.46e-3 & 8.75e-3 & 3.01e-2 & 2.71e-1 & 1.17e+00 \\
$k$-NN$^\ddagger$ & 3.33e-3 & 4.35e-3 & 7.69e-3 & 1.60e-2 & 6.67e-2 & 1.82e-1 \\
\multicolumn{7}{l}{\textit{T3, largest $n$ per $d$}} \\
BLML, $f_c=0.7$ & 2.09e-5 & 2.53e-4 & -- & 2.58e-3 & -- & \textbf{1.27e-2} \\
BLML, $f_c=1$ & \textbf{2.08e-5} & \textbf{8.50e-5} & -- & \textbf{8.38e-4} & -- & 3.50e-2 \\
BLML, $f_c=2$ & 4.24e-5 & 4.06e-4 & \textbf{2.42e-3} & 1.36e-2 & 3.10e-1 & 1.42e+00 \\
BLML, $f_c=4$ & 1.24e-4 & 1.79e-3 & 2.03e-2 & 2.17e-1 & 3.68e+00 & 1.15e+02 \\
KDE, oracle $h$ & 9.71e-5 & 8.11e-4 & 2.98e-3 & 1.00e-2 & \textbf{4.57e-2} & 1.25e-1 \\
KDE, LSCV $h$ & 1.02e-4 & 8.31e-4 & 2.98e-3 & 1.00e-2 & \textbf{4.57e-2} & 1.25e-1 \\
KDE, Scott $h$ & 1.00e-4 & 8.70e-4 & 3.58e-3 & 1.31e-2 & 6.31e-2 & 1.71e-1 \\
6th-order KDE, oracle $h$$^\dagger$ & 2.87e-5 & 2.21e-4 & 6.57e-4 & 2.16e-3 & 1.13e-2 & 4.45e-2 \\
sinc KDE, $f_c=4$$^\dagger$ & 1.27e-4 & 1.80e-3 & 2.02e-2 & 2.25e-1 & 2.66e+01 & 3.02e+03 \\
Adaptive KDE & 1.31e-4 & 1.10e-3 & 4.77e-3 & 1.69e-2 & 2.06e-1 & 1.18e+00 \\
$k$-NN$^\ddagger$ & 3.17e-3 & 3.53e-3 & 4.69e-3 & 9.48e-3 & 4.42e-2 & 1.38e-1 \\
\bottomrule
\end{tabular}%
}
\end{table}

\subsection{Kinks and edges: where the band limit stops paying}
\label{sec:exp-nonsmooth}

The Gaussian has a spectrum that decays faster than any power, so a moderate cutoff leaves almost no bias. Two
further truths have spectra that decay slowly: a product of Laplace densities, $f=\prod_k\frac12e^{-|x_k|}$, whose
square root has a kink at the mode, and a product of Beta$(2,2)$ densities on $[-3,3]$, whose support has hard
edges. Neither has a true cutoff, so here every estimator chooses its smoothing from the data: BLML by the
validation search of the selector (and, for comparison, by the streaming selector of
Section~\ref{sec:grad}), the Gaussian KDE by Scott's rule, least-squares or likelihood cross-validation, and
Abramson's adaptive KDE with a likelihood-cross-validated pilot. Table~\ref{tab:nonsmooth} gives the relative ISE
for $d\in\{2,4,8\}$ and $n\in\{10^3,10^4\}$, medians of four replicates.

\begin{table}[t]
\centering\small
\caption{Truths that are not smooth: relative ISE (median of four replicates) with every estimator's smoothing
chosen from the data. Best per row in bold; the best KDE's bandwidth rule in parentheses.}
\label{tab:nonsmooth}
\begin{tabular}{rrrrl}
\toprule
$d$ & $n$ & BLML, one-shot & BLML, streaming & best data-driven KDE \\
\midrule
\multicolumn{5}{l}{\textit{Laplace (kink)}} \\
2 & $10^{3}$ & 0.0486 & 0.0776 & \textbf{0.0213} (adaptive) \\
2 & $10^{4}$ & 0.0216 & 0.0215 & \textbf{0.00599} (adaptive) \\
4 & $10^{3}$ & 0.279 & 0.266 & \textbf{0.0972} (adaptive) \\
4 & $10^{4}$ & 0.141 & 0.147 & \textbf{0.0405} (adaptive) \\
8 & $10^{3}$ & 0.66 & 0.74 & \textbf{0.545} (LSCV) \\
8 & $10^{4}$ & 0.499 & 0.603 & \textbf{0.384} (LSCV) \\
\multicolumn{5}{l}{\textit{bounded (edges)}} \\
2 & $10^{3}$ & 0.0295 & \textbf{0.017} & 0.0179 (Scott) \\
2 & $10^{4}$ & 0.00619 & 0.00511 & \textbf{0.00376} (Scott) \\
4 & $10^{3}$ & 0.0697 & 0.0779 & \textbf{0.0687} (Scott) \\
4 & $10^{4}$ & 0.0263 & \textbf{0.0221} & 0.0269 (Scott) \\
8 & $10^{3}$ & 0.123 & \textbf{0.114} & 0.301 (LSCV) \\
8 & $10^{4}$ & \textbf{0.136} & 0.17 & 0.184 (Scott) \\
\bottomrule
\end{tabular}
\end{table}

The kink defeats BLML. On the Laplace truth the adaptive KDE, whose kernels narrow at the peak, is 1.2--3.6 times
more accurate in every cell, and BLML wins none of the \NsCells. A kink keeps the spectrum heavy, a band-limited fit
answers it with ripples that spread away from the peak, and one global band cannot be sharp at the mode and smooth
in the tails at once. The edges are a different matter: BLML is close to the best KDE in two dimensions and ahead
in \NsWinsB\ of the \NsCells\ cells, by 2.6 times at $d=8$, $n=10^3$, where the kernel estimators degrade faster than
BLML does. So the band limit pays for smooth densities, still pays in high dimension for densities with edges, and
does not pay for densities with a kink, where a locally adaptive estimator should be used.

\subsection{Against a normalizing flow}
\label{sec:exp-flow}

Neural density estimators are the modern alternative in moderate dimension. They are flexible parametric models in
the sense above, so the comparison is one of accuracy, not of like estimators. We fit a neural spline flow
\citep{durkan2019neural,papamakarios2021normalizing} (five rational-quadratic spline transforms, two hidden layers of
128 units, Adam, early stopping on a held-out fifth of the sample, data standardised per dimension) to the same
truths and score it in the same way; BLML chooses its cut-off by the validation search, so both estimators are
fully data-driven (Table~\ref{tab:flow}, medians of two to four replicates).

\begin{table}[t]
\centering\small
\caption{Relative ISE of BLML (cut-off by validation search) and the best data-driven KDE, both nonparametric,
and a neural spline flow, a flexible parametric reference (medians over replicates; best per row in bold).}
\label{tab:flow}
\begin{tabular}{rrrrr}
\toprule
$d$ & $n$ & BLML & neural spline flow & best KDE \\
\midrule
\multicolumn{5}{l}{\textit{band-limited, separable (T1)}} \\
2 & $10^{3}$ & \textbf{0.0049} & 0.034 & 0.013 \\
2 & $10^{4}$ & \textbf{0.00055} & 0.0072 & 0.0029 \\
4 & $10^{3}$ & \textbf{0.029} & 0.065 & 0.064 \\
4 & $10^{4}$ & \textbf{0.00089} & 0.016 & 0.018 \\
8 & $10^{3}$ & 0.15 & \textbf{0.14} & 0.33 \\
8 & $10^{4}$ & \textbf{0.017} & 0.031 & 0.18 \\
8 & $10^{5}$ & \textbf{0.0013} & 0.01 & 0.093 \\
\multicolumn{5}{l}{\textit{correlated normal (T3)}} \\
2 & $10^{3}$ & \textbf{0.0035} & 0.033 & 0.013 \\
2 & $10^{4}$ & \textbf{0.00092} & 0.0091 & 0.0028 \\
4 & $10^{3}$ & \textbf{0.022} & 0.049 & 0.078 \\
4 & $10^{4}$ & \textbf{0.0043} & 0.016 & 0.023 \\
8 & $10^{3}$ & 0.18 & \textbf{0.15} & 0.39 \\
8 & $10^{4}$ & 0.063 & \textbf{0.042} & 0.22 \\
8 & $10^{5}$ & 0.031 & \textbf{0.014} & 0.13 \\
\multicolumn{5}{l}{\textit{Laplace (kink)}} \\
2 & $10^{3}$ & 0.049 & 0.047 & \textbf{0.021} \\
2 & $10^{4}$ & 0.022 & 0.008 & \textbf{0.006} \\
4 & $10^{3}$ & 0.28 & \textbf{0.063} & 0.097 \\
4 & $10^{4}$ & 0.14 & \textbf{0.014} & 0.041 \\
8 & $10^{3}$ & 0.66 & \textbf{0.15} & 0.54 \\
8 & $10^{4}$ & 0.5 & \textbf{0.042} & 0.38 \\
\multicolumn{5}{l}{\textit{bounded (edges)}} \\
2 & $10^{3}$ & 0.029 & 0.044 & \textbf{0.018} \\
2 & $10^{4}$ & 0.0062 & 0.0078 & \textbf{0.0038} \\
4 & $10^{3}$ & 0.07 & 0.1 & \textbf{0.069} \\
4 & $10^{4}$ & 0.026 & \textbf{0.025} & 0.027 \\
8 & $10^{3}$ & \textbf{0.12} & 0.14 & 0.3 \\
8 & $10^{4}$ & 0.14 & \textbf{0.066} & 0.18 \\
\bottomrule
\end{tabular}
\end{table}

Where the band limit holds, BLML is far ahead. On the band-limited truth it has the lowest error in six of seven
cells, by a factor of 2 to 18 once $n\ge10^4$ (at $d=4$, $n=10^4$: $0.0009$ against $0.016$); only at $d=8$ with
$10^3$ points does the flow match it. On the Gaussian, BLML is 4 to 10 times more accurate in two and four
dimensions, and the flow is 1.2 to 2.2 times more accurate in eight. On the kink the flow is the best estimator in
high dimension by a wide margin ($0.04$ against $0.50$ for BLML and $0.38$ for the best KDE at $d=8$, $n=10^4$): a
flow bends space locally, which a single global band cannot. With edges the flow wins only at $d=8$, $n=10^4$. The
pattern is the one the rate analysis predicts: the band limit buys an order of magnitude when it describes the
density, and nothing when the density has local structure the band cannot resolve.

The advantage is not an artefact of one flow size: flows with 868 and 8{,}504 weights fare no better
(Appendix~\ref{app:flow-sizes}).

\subsection{Solver behavior}
\label{sec:exp-solver}

\paragraph{Convergence.}
Every fit we report converged to $\max_b|\rho_b|<10^{-4}$ with a monotonically decreasing
objective, in 9--11 sweeps regardless of $n$ or $d$ (6--11 for $d=1$ to $16$); without the descent
safeguard the one-dimensional fit needed \UnsafeSweepsDOne\ sweeps. Convergence traces are in
Appendix~\ref{app:solver}.

\paragraph{Block size.}
Larger blocks amortize kernel launches: $B_s=512$ fits twelve times faster than $B_s=32$ with the
same result (Appendix Table~\ref{tab:block}).

\paragraph{Precision.}
float32 and float64 run at comparable speed and agree to within the solver tolerance
(Appendix~\ref{app:solver}).

\paragraph{CPU versus GPU.}
Against the same algorithm compiled with Numba and run single-threaded on the CPU, the GPU
implementation fits \SpeedFitRange\ faster (the speedup grows with $n$) and predicts
\SpeedPredRange\ faster, with test log-likelihoods that agree to $3\times10^{-5}$
(Appendix Table~\ref{tab:speed}).
\section{Experiments: learning the cut-off}
\label{sec:exp}

Does the streaming fit deliver what Section~\ref{sec:learn} promises: the exact fit, the right cut-off, and a saving
in time? We test each in turn.

\subsection{Convergence at a fixed cut-off}
\label{sec:fixed}

Before it learns anything, streaming must reach the exact fit. At a fixed cut-off ($0.4$ for T1 and T2, $1.0$ for
T3) we run it and the solver to the tolerance $10^{-4}$ at $d=4$, $n\in\{2\cdot10^4,5\cdot10^4\}$, three replicates,
in float64 on a laptop CPU, streaming in batches of $n/20$.

\begin{figure}[t]
  \centering
  \includegraphics[width=\textwidth]{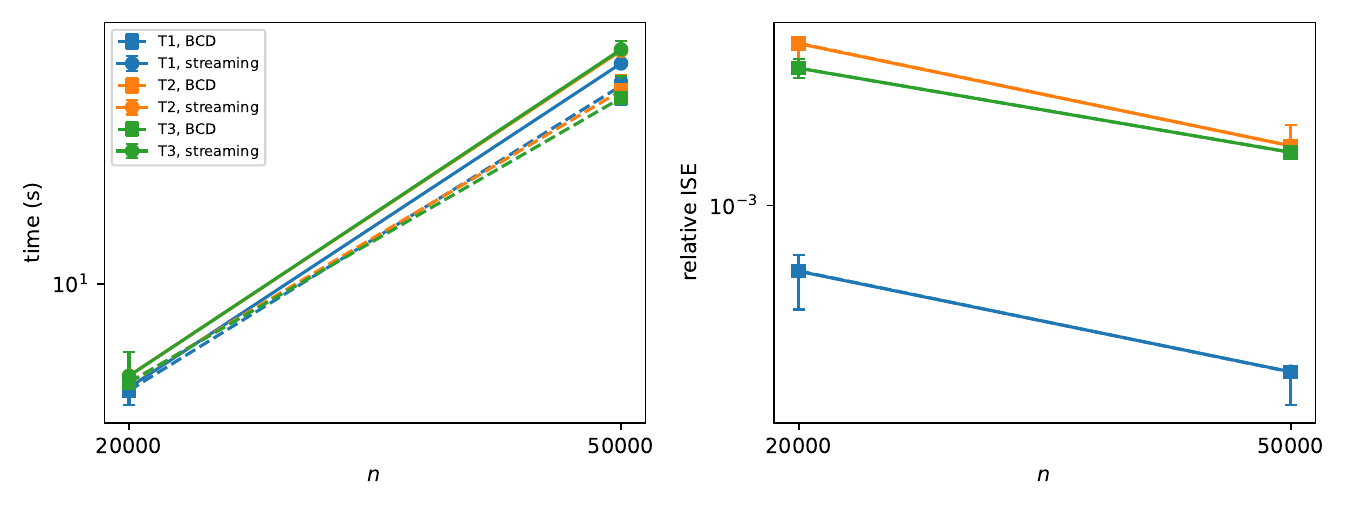}
  \caption{Fixed cut-off, $d=4$, both to tolerance $10^{-4}$ (CPU, float64). Left: fit time; right:
  relative ISE. Medians over three replicates, bars min--max. Dashed: BCD; solid: streaming.}
  \label{fig:fixed}
\end{figure}

It does: the ratio of its error to BCD's is
\FixedIseRatioRange\ in every cell (Fig.~\ref{fig:fixed}). It needs \FixedEpochs\ epochs where BCD takes
\FixedSweeps\ sweeps, so an epoch does the work of a sweep, and despite the per-batch overhead (the touched
rows are computed batch by batch rather than in one pass) it takes \FixedSpeedTwentyK\ times BCD's time at
$2\cdot10^4$ points and \FixedSpeedFiftyK\ times at $5\cdot10^4$.

\subsection{Cut-off recovery on synthetic truths}
\label{sec:synth}

Does the stream find the right cut-off? On T1, T2 and T3 at $d\in\{2,4,8\}$, $n\in\{10^3,10^4\}$ (four
replicates), streaming BLML sees the points in random order, with a burn-in on $\min(n/2,5000)$ points, batches of
$n/20$ and two passes; we compare it with BLML at the true cut-off and with Gaussian KDEs under four data-driven
bandwidths (Scott's rule, least-squares and likelihood cross-validation, Abramson's adaptive rule).

\begin{figure}[t]
  \centering
  \includegraphics[width=\textwidth]{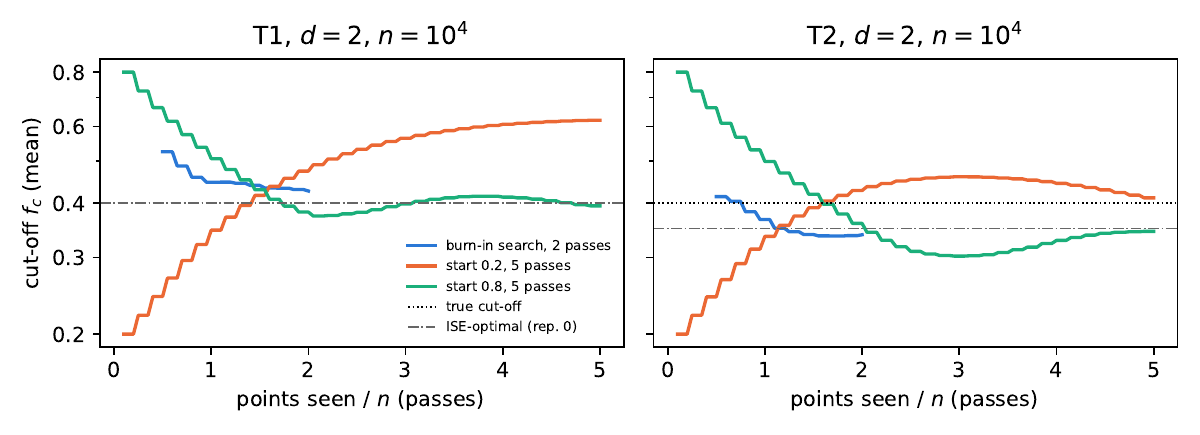}
  \caption{Cut-offs of streaming BLML against points seen (in passes over the data), two-dimensional
  band-limited truths, $n=10^4$; geometric mean over four replicates and both columns. Dotted: the true
  cut-off; dash-dotted: the common cut-off of least integrated squared error on replicate 0.}
  \label{fig:trace}
\end{figure}

Figure~\ref{fig:trace} follows the cut-offs on the two-dimensional
band-limited truths at $n=10^4$ (batches of $n/20$, four replicates), from the burn-in search and from fixed
starts at half and twice the true cut-off. From above, the cut-offs come down to the true one (T1; relative
rms error $\TraceTOneErrHigh\%$ after five passes) or to the error-minimising one (T2, where it lies below
$0.4$ at this $n$). From below they rise steeply and overshoot on T1, to about $0.6$ (error
$\TraceTOneErrLow\%$): the held-out likelihood is steep below the true cut-off and nearly flat above it
(0.002 nats per point between $0.4$ and $0.6$, while the integrated squared error grows $2.6$-fold), so the
gradient that drives them up is strong and the one that brings them back is weak.

\begin{figure}[t]
  \centering
  \includegraphics[width=\textwidth]{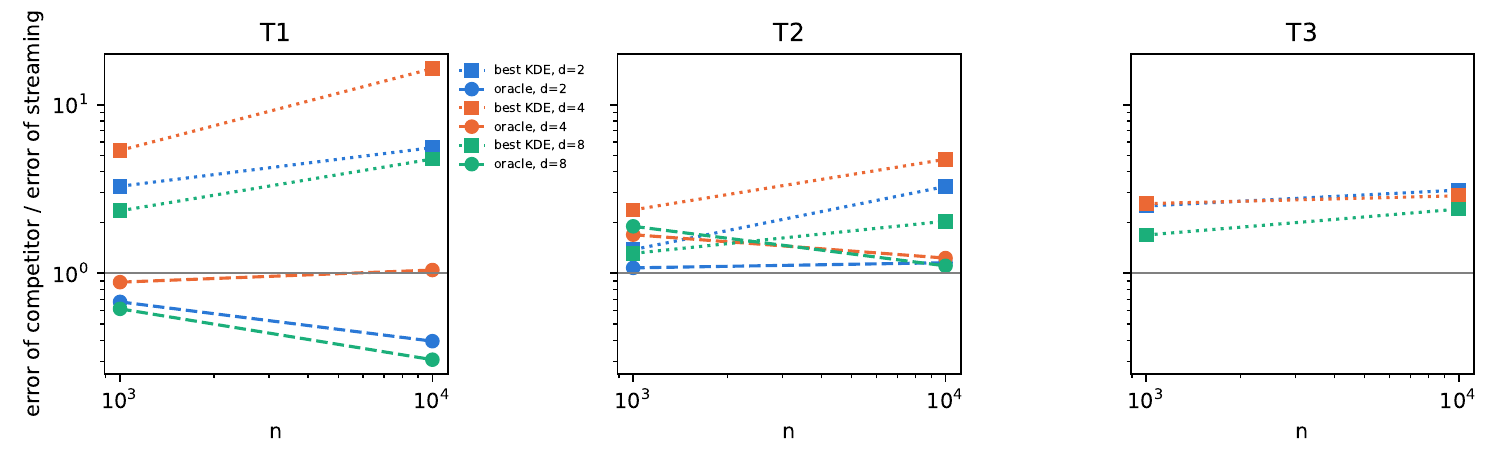}
  \caption{Error of a competitor divided by the error of streaming BLML (relative ISE, median over
  replicates); above 1, streaming is better. Dotted: the best of four data-driven Gaussian KDEs in that
  cell; dashed: BLML at the true cut-off (oracle; T1, T2), which sits near 1.}
  \label{fig:synth}
\end{figure}

\begin{table}[t]
  \centering\small
  \caption{Streaming BLML against the truth. $f_c$ error: relative rms error of the learned cut-offs,
  $\bigl(\frac1d\sum_j(f_{c,j}/f_c^{\text{true}}-1)^2\bigr)^{1/2}$ (T1, T2; T3 has no true cut-off).
  Relative ISE of streaming BLML, the oracle and the best data-driven Gaussian KDE (rule in parentheses).
  Medians over replicates.}
  \label{tab:synth}
  \begin{tabular}{llrcccc}
\toprule
& & & $f_c$ error & \multicolumn{3}{c}{relative ISE} \\
\cmidrule(lr){5-7}
truth & $d$ & $n$ & (rel.\ rms) & streaming & oracle & best KDE \\
\midrule
T1 & 2 & $10^{3}$ & 12\% & $4.08\text{e-}3$ & $2.75\text{e-}3$ & $1.34\text{e-}2$ (Scott) \\
T1 & 2 & $10^{4}$ & 9\% & $5.24\text{e-}4$ & $2.07\text{e-}4$ & $2.91\text{e-}3$ (Scott) \\
T1 & 4 & $10^{3}$ & 16\% & $1.19\text{e-}2$ & $1.05\text{e-}2$ & $6.38\text{e-}2$ (Scott) \\
T1 & 4 & $10^{4}$ & 7\% & $1.12\text{e-}3$ & $1.17\text{e-}3$ & $1.83\text{e-}2$ (adaptive) \\
T1 & 8 & $10^{3}$ & 26\% & $1.42\text{e-}1$ & $8.70\text{e-}2$ & $3.32\text{e-}1$ (Scott) \\
T1 & 8 & $10^{4}$ & 19\% & $3.71\text{e-}2$ & $1.14\text{e-}2$ & $1.76\text{e-}1$ (Scott) \\
T2 & 2 & $10^{3}$ & 29\% & $2.13\text{e-}2$ & $2.30\text{e-}2$ & $2.94\text{e-}2$ (adaptive) \\
T2 & 2 & $10^{4}$ & 16\% & $1.68\text{e-}3$ & $1.94\text{e-}3$ & $5.49\text{e-}3$ (adaptive) \\
T2 & 4 & $10^{3}$ & 35\% & $5.06\text{e-}2$ & $8.56\text{e-}2$ & $1.20\text{e-}1$ (adaptive) \\
T2 & 4 & $10^{4}$ & 26\% & $8.21\text{e-}3$ & $1.01\text{e-}2$ & $3.89\text{e-}2$ (adaptive) \\
T2 & 8 & $10^{3}$ & 43\% & $2.99\text{e-}1$ & $5.68\text{e-}1$ & $3.92\text{e-}1$ (adaptive) \\
T2 & 8 & $10^{4}$ & 37\% & $1.27\text{e-}1$ & $1.40\text{e-}1$ & $2.58\text{e-}1$ (LSCV) \\
T3 & 2 & $10^{3}$ & -- & $5.03\text{e-}3$ & -- & $1.26\text{e-}2$ (LSCV) \\
T3 & 2 & $10^{4}$ & -- & $9.16\text{e-}4$ & -- & $2.85\text{e-}3$ (LSCV) \\
T3 & 4 & $10^{3}$ & -- & $3.03\text{e-}2$ & -- & $7.84\text{e-}2$ (LSCV) \\
T3 & 4 & $10^{4}$ & -- & $7.89\text{e-}3$ & -- & $2.27\text{e-}2$ (adaptive) \\
T3 & 8 & $10^{3}$ & -- & $2.29\text{e-}1$ & -- & $3.85\text{e-}1$ (LSCV) \\
T3 & 8 & $10^{4}$ & -- & $9.15\text{e-}2$ & -- & $2.20\text{e-}1$ (LSCV) \\
\bottomrule
\end{tabular}
\end{table}

And the accuracy follows: streaming BLML has lower error than the best data-driven kernel estimator in
\SynStreamBeatsKDE\ of \SynStreamCells\ cells (error ratios \SynKDEOverStreamRange;
Fig.~\ref{fig:synth}, Table~\ref{tab:synth}), and its error is a median factor \SynStreamOverOracleMed\ of the oracle's
(range \SynStreamOverOracleRange). The learned cut-offs are within \SynFcErrRange\% of the true cut-off
(relative rms over dimensions, median \SynFcErrMed\%), and the error falls with $n$ in every cell. The
remaining deviation is mostly downward and largest for the non-separable T2 at $d\ge4$; there the
lower cut-off trades a little bias for less variance, and streaming's error is at or below the
oracle's, so at these sample sizes $f_c^{\text{true}}$ is not the error-minimising cut-off. The largest
gap to the oracle is T1 at $d=8$, $n=10^4$ (a factor 3.3), where the cut-offs are 19\% low: the burn-in
selects them on half the sample and the remaining batches allow few Adam steps; six passes instead of two
change the errors at $d=8$ by less than their spread over replicates.

\subsection{Speed against one-shot selection}
\label{sec:speed}

How much time does learning the cut-off on the way save? The benchmark is a two-dimensional band-limited mixture of
three products of $\mathrm{sinc}^2$ factors (heavy tails), with $2000$ points held out. Streaming uses batches of 2000
(CPU) or 4000 (GPU) points after a burn-in of $5\cdot10^3$ or $10^4$; one-shot selection searches on all points and
refits. All runs are in float32, on a laptop CPU or a T4.

\begin{table}[t]
\centering
\caption{GPU (T4): streaming BLML (two passes) against one-shot selection, two-dimensional benchmark.}
\label{tab:gpu}
\fitwidth{%
\begin{tabular}{rrrrrrr}
\toprule
& \multicolumn{2}{c}{streaming (s)} & one-shot & & \multicolumn{2}{c}{held-out LL} \\
\cmidrule(lr){2-3}\cmidrule(lr){6-7}
$n$ & total & burn-in & (s) & speed-up & streaming & one-shot \\
\midrule
$20$K & 11.9 & 8.1 & 15.3 & 1.3$\times$ & -2.4143 & -2.4214 \\
$40$K & 14.1 & 4.0 & 35.0 & 2.5$\times$ & -2.2218 & -2.2247 \\
$100$K & 41.9 & 5.0 & 175.7 & 4.2$\times$ & -2.2575 & -2.2600 \\
\bottomrule
\end{tabular}%
}
\end{table}

\begin{table}[t]
\centering
\caption{CPU (laptop): streaming BLML (two passes) against one-shot selection, two-dimensional benchmark.}
\label{tab:cpu}
\fitwidth{%
\begin{tabular}{rrrrrrr}
\toprule
& \multicolumn{2}{c}{streaming (s)} & one-shot & & \multicolumn{2}{c}{held-out LL} \\
\cmidrule(lr){2-3}\cmidrule(lr){6-7}
$n$ & total & burn-in & (s) & speed-up & streaming & one-shot \\
\midrule
$20$K & 16.6 & 7.1 & 73.4 & 4.4$\times$ & -2.4257 & -2.4214 \\
$40$K & 50.0 & 6.9 & 254.4 & 5.1$\times$ & -2.2262 & -2.2247 \\
\bottomrule
\end{tabular}%
}
\end{table}

On the CPU streaming is \CpuSpeedupRange$\times$ faster (Table~\ref{tab:cpu}), and the gain grows with $n$ as the
cost analysis predicts: one-shot selection pays tens of fits on all the data, streaming about two sweeps per pass plus
a burn-in of fixed size. Its held-out log-likelihood is within $0.005$ nats per point of one-shot selection's, inside
the spread over seeds ($0.015$). On the GPU (Table~\ref{tab:gpu}) it is \GpuSpeedupRange$\times$ faster and
$0.003$--$0.007$ nats per point better. The GPU gain is smaller because there one-shot selection is already fast and
the burn-in search, a few seconds whatever its size, dominates at $2\cdot10^4$ points.
\newcommand{\KdeOracleSlopeLo}{$-0.85$}
\newcommand{\KdeOracleSlopeHi}{$-0.77$}
\newcommand{\BlmlSlopeLo}{$-2.32$}
\newcommand{\BlmlSlopeHi}{$-1.19$}
\newcommand{\HOracleSlopeLo}{$-0.27$}
\newcommand{\HOracleSlopeHi}{$-0.10$}
\newcommand{\HStreamTOneLo}{$+0.14$}
\newcommand{\HStreamTOneHi}{$+0.20$}
\newcommand{\StreamKdeTOneIseLo}{$+0.36$}
\newcommand{\StreamKdeTOneIseHi}{$+0.41$}
\newcommand{\KdeBlmlIseWins}{5}
\subsection{The same learning applied to a KDE}
\label{sec:kde}

The oracle comparisons of Section~\ref{sec:experiments} already show that BLML's lead is not a matter of tuning:
no bandwidth gives a KDE BLML's rate. What they do not show is whether the streaming machinery is specific to BLML.
We therefore give a Gaussian KDE the same learning: bandwidths by Adam on the same prequential score, with the same
burn-in, batches and learning rate, on T1, T2 and T3 in one and two dimensions up to $n=64{,}000$ (two seeds),
against batch KDEs at Scott's rule, likelihood cross-validation and the oracle bandwidth.

The learning runs, but it cannot change the rate (Appendix Tables~\ref{tab:kde_ise} and~\ref{tab:kde_kl}). Streaming
KDE's learned bandwidth lands within a factor of three of the oracle's (0.7 to 2.8), and even the oracle KDE's ISE
falls only with slope \KdeOracleSlopeLo{} to \KdeOracleSlopeHi{} on T1 and T2, near the
theoretical $-0.8$; streaming BLML's falls with slope \BlmlSlopeLo{} to \BlmlSlopeHi{} and is lowest at $n=64{,}000$
in \KdeBlmlIseWins{} of six cases (it ties the oracle KDE in the sixth). One failure is the KDE's own: on the
heavy-tailed T1 its learned bandwidth drifts to about twice the oracle's and its error rises with $n$, because the
log score is dominated by sparse tail points, which tend to reward a wider kernel, and one bandwidth must trade the
centre for the tails. A band limit constrains the shape of $\sqrt f$, not a scale tied to local density, and BLML's
learned cut-off shows no such drift.
\section{Real data: electricity prices}
\label{sec:real-prices}

Does any of this survive data the band limit does not describe, with a cut-off nobody knows? Hourly electricity
prices are heavy-tailed, multimodal and periodic in the hour of the day. We estimate the conditional density of the
hourly day-ahead locational marginal price $y$ at CAISO's Northern California trading hub
(TH\_NP15\_GEN-APND; 2023-06-29 to 2024-11-30, 12{,}502 hours, $-\$30.7$ to $\$487.4$ per MWh, 2.5\% of hours negative) given the hour and the price 24 hours earlier, and given the hour and the
prices 24 and 168 hours earlier: the covariates of a day-ahead forecaster.
Quantised variables are jittered by one quantisation step.

\label{sec:prices}
\paragraph{Protocol.} A random 20\% of calendar days ($\DayNdays$ days, $\DayNtest$ hours) is held out for testing.
Holding out whole days rather than hours stops a model from interpolating a test hour from its neighbours in the
same day; with random hours, one of the baselines gained up to $0.8$ nats per hour that way. Every model is tuned on
the same validation set, a random 20\% of the training days, and refitted on all training days. The score is the
mean test log-likelihood in nats per hour.

\paragraph{BLML.} The conditional density is the ratio of a joint fit, three-dimensional (price, hour, one lag) or
four-dimensional (price, hour, two lags), to a fit of the covariates alone with the
same cut-offs, computed in one fused pass (Appendix~\ref{app:prices} gives the construction and why its score is conservative). The hour is a Dirichlet column with period 24. All cut-offs are learned
by the streaming selector of this paper: an exact fit with a validation search on a burn-in set that holds out
whole days, then batches of 42 days, each scored by the current model before it is used, for six passes. The hour
moves in whole harmonics. On a laptop CPU this takes 53--92\,s for the $\approx10^4$ training hours.

\paragraph{Baselines.} The nonparametric baselines assume, like BLML, no parametric form: a conditional KDE
\citep{hyndman1996conditional} with a von Mises kernel on the hour and all bandwidths chosen on validation, and a
conditional $k$-nearest-neighbour estimator with $k$ and its price bandwidth chosen on validation. As a reference
for what a well-matched inductive bias achieves we also fit structured models: a Gaussian mixture on the joint
space, XGBoostLSS with a Gaussian output \citep{marz2019xgboostlss,chen2016xgboost}, a mixture density network
\citep{bishop1994mdn} and TabPFN \citep{hollmann2025tabpfn}, a transformer pretrained on synthetic tables that
predicts from the training data as context without being fitted to it.

\begin{table}[t]
  \centering\small
  \caption{Test log-likelihood of the CAISO price on held-out days (nats per hour, higher is better; best per group
  in bold). Columns are the covariate sets. $^*$TabPFN is pretrained on synthetic tables and is not fitted to these
  data.}
  \label{tab:prices}
  \begin{tabular}{lrr}
\toprule
model & hour, lag 24\,h & hour, lags 24, 168\,h \\
\midrule
\multicolumn{3}{l}{\emph{nonparametric (complexity grows with $n$)}} \\
BLML (this paper) & \textbf{-3.393} & \textbf{-3.383} \\
conditional KDE & -3.787 & -3.850 \\
conditional $k$-NN & -3.703 & -3.654 \\
\midrule
\multicolumn{3}{l}{\emph{flexible parametric, semiparametric and neural (fixed architecture; reference)}} \\
Gaussian mixture & -3.425 & -3.427 \\
XGBoostLSS (Gaussian) & -3.524 & -3.692 \\
mixture density network & -3.362 & -3.426 \\
neural spline flow & -3.440 & -3.501 \\
TabPFN$^*$ & \textbf{-3.334} & \textbf{-3.367} \\
\bottomrule
\end{tabular}
\end{table}

\paragraph{Results.} Among the nonparametric estimators BLML scores best on both covariate sets (Table~\ref{tab:prices}). With the
hour and yesterday's price it scores $\DayStreamHL$ nats per hour, $\DayStreamVsCondKNNHL$ above the conditional
$k$-NN and $\DayStreamVsCondKDEHL$ above the conditional KDE; with both lags the margins are
$\DayStreamVsCondKNNHLL$ and $\DayStreamVsCondKDEHLL$. Given yesterday's price, today's price lies in a narrow band,
and the kernel estimators' bandwidths, chosen on the same validation days, cannot follow it. BLML is also above the
Gaussian mixture, XGBoostLSS and a conditional neural spline flow on both covariate sets (by $0.047$ and $0.118$ nats
per hour over the flow), and with both lags above the mixture density network. With
yesterday's price alone the mixture density network is ahead by $\DayStreamVsMDNHLAbs$, and TabPFN, which brings a
prior learned in pretraining, leads on both sets (by $\DayStreamVsTabPFNHLAbs$ and $\DayStreamVsTabPFNHLLAbs$).

Streaming matches one-shot selection of the same model, which refits on all training days at every candidate: with
yesterday's price one-shot selection scores $-3.394$ nats per hour against streaming's $-3.393$, and with both lags
$-3.424$ against $-3.383$; a second streaming seed gives $-3.397$ and $-3.400$. The learned cut-offs also do
variable selection. Adding last week's price changes BLML's score by
$\DayAddWeekBLMLstream$ nats per hour, within the spread over burn-in draws, because the cut-off on that column
falls to $\DayStreamFcLagWeek$: the band is nearly flat along it and the covariate is effectively dropped. One-shot selection keeps it, which is where its
$0.04$ deficit with both lags comes from. The mixture density network, TabPFN and the KDE lose $0.03$--$0.06$ nats per hour when it is added, and XGBoostLSS
loses $0.17$.
\section{Standard benchmarks: POWER and GAS}
\label{sec:uci}

Neural density estimators are compared on a common suite of tabular data sets \citep{papamakarios2017maf}. Two of
them fall in our setting, dense data in at most eight dimensions, and have enough points to test a nonparametric
estimator at full size: POWER (household electricity consumption, $d=6$, $1.66\times10^6$ training points) and GAS
(chemical sensor readings, $d=8$, $8.5\times10^5$). We use the published preprocessing and splits, so our scores
compare directly with the published ones; the other sets of the suite have 21 to 63 dimensions, beyond the range we study
(Section~\ref{sec:rate}).

\paragraph{Protocol.} BLML is fitted by streaming (Section~\ref{sec:stream}) to the whole training split, in batches
of $10^4$ points for three passes, float32 on one A100 GPU; the cut-off starts from the end of a streaming run on a
$10^5$-point subsample. The Gaussian product KDE uses all training points, with the multiplier of Scott's rule
chosen on the validation split over $[0.01, 2]$. The neural scores are those published with the suite
\citep{germain2015made,dinh2017realnvp,papamakarios2017maf,durkan2019nsf}; each architecture was tuned on these data.

\begin{table}[t]
  \centering
  \caption{Test log-likelihood on the standard benchmarks (nats per point, higher is better; full training splits,
  published preprocessing). Best per group in bold.}
  \label{tab:uci}
  \begin{tabular}{lrr}
\toprule
model & POWER ($d=6$) & GAS ($d=8$) \\
\midrule
\multicolumn{3}{l}{\emph{nonparametric (complexity grows with $n$)}} \\
BLML, streaming cut-off (this paper) & \textbf{0.06} & \textbf{12.51} \\
Gaussian KDE, bandwidth on validation & $-0.33$ & 10.93 \\
\midrule
\multicolumn{3}{l}{\emph{neural (fixed architecture; published, reference)}} \\
MADE & $-3.08$ & 3.56 \\
Real NVP & $-0.02$ & 4.78 \\
MAF & 0.24 & 10.08 \\
neural spline flow & \textbf{0.66} & \textbf{13.09} \\
\bottomrule
\end{tabular}
\end{table}

\paragraph{Results.} BLML is the better nonparametric estimator on both sets, by $0.39$ nats per point on POWER
and $1.58$ on GAS (Table~\ref{tab:uci}), and it is ahead of Real NVP on both and of MAF on GAS; the neural spline
flow is ahead by $0.60$ and $0.58$. We have not found nonparametric scores reported on the suite before. The fits take 79 minutes for POWER and 38 for GAS, and the held-out score has settled after one or two passes
(POWER $0.067$, $0.072$, $0.070$; GAS $12.518$, $12.514$, $12.512$ on a validation subsample). On $10^5$-point
subsamples, streaming matched one-shot selection with the cut-off rescaled to $n$ ($-0.203$ against $-0.205$ on
POWER) or beat it ($11.26$ against $11.15$ on GAS), where it raised one cut-off from $7.6$ to $72$, a value the
search never reached. GAS is concentrated: its readings follow a slow trajectory, and the median
distance to a nearest neighbour is $0.03$ standard deviations, against $0.7$ for Gaussian data of the same size and
dimension. The learned cut-offs follow it, rising to $20$--$90$ on most axes, where those for POWER stay below $10$.
\section{Discussion}
\label{sec:discussion}

\paragraph{What we learned.} A nonparametric estimator can keep the parametric rate as the dimension grows: on
band-limited truths BLML's error falls as $n^{-1}$ in every dimension we tested, up to eight. The sinc-kernel
estimator, linear in the data and unbiased when its cut-off covers the band, attains $n^{-1}$ too, but it is not a
density: it goes negative at up to half of the sample points. BLML keeps the rate and returns
a proper density, and it does so with a smaller constant: it learns $\sqrt f$, whose band is half that of $f$, so it
needs about $2^{-d}$ as many effective parameters. The estimator can also choose its own smoothing: a
streaming fit learns the cut-off for about the cost of one more fit, and a kernel estimator given the same learning
keeps its slower rate. Both hold up on data the assumption does not describe: on electricity prices BLML leads every
other nonparametric estimator we tested and a neural spline flow, and on the standard POWER and GAS benchmarks it leads a tuned
KDE at full size and sits between early and recent flows, with no architecture to choose.

\paragraph{A loss is informative.} BLML's assumption is a single stated property, so when it loses to a flexible
model on ample data, the loss says something about the truth: it is not band-limited at any band the data support. A
neural model's win or loss says little, because its inductive bias is implicit. The learned cut-off sharpens the
check: for a band-limited truth the best cut-off stops moving once $n$ is large, and for one that is not it keeps
rising (Fig.~\ref{fig:fc-sweep}). Whether the cut-off settles as data accumulate is thus a test of the model class,
though a weak one at small $n$ in high dimension, where BLML's constant rather than its assumption sets its error. The
losses we found fit this reading. On a density with a kink an adaptive KDE and a flow beat BLML in every cell, by up to
an order of magnitude in high dimension: a single global band cannot be sharp at the peak and smooth in the tails. With
hard edges BLML is close in low dimension and ahead in high dimension.

\paragraph{Fixed architectures.} BLML's effective complexity is set by its learned cut-off and grows with the data; a
model of fixed architecture has to be resized as the data grow. Modern networks can be made large enough that this
rarely matters in practice, and where the density has structure a band cannot encode, a well-chosen network wins: on
the kink, and on POWER and GAS, where the best published flow is about $0.6$ nats per point ahead after its
architecture was tuned on those data. A product band is one cut-off per axis; a flow can bend its coordinates to
the density. What BLML offers in exchange is no architecture to choose, a single stated assumption whose failure
the learned cut-off reveals, and a fit whose cost and memory are known in advance.

\paragraph{Limitations.} The rate is measured, not proved: the effective-dimension argument explains the exponent
and the constant, but extending the consistency proof of \citet{agarwal2017blml} to a rate is open; Appendix~\ref{app:conjecture} states the
conjecture and the three steps a proof needs. The cost is
quadratic in the number of bins once binning stops compressing: seconds at $n=10^5$ and ten minutes at $10^6$ on one
A100 GPU, but beyond that the solver must be distributed (its blocks partition naturally across devices) or the expansion
approximated. The product kernel's band is an axis-aligned box. The benchmark
suite's other sets, with 21 to 63 dimensions, are outside the range we study: with the constant of Section~\ref{sec:rate}, an
estimator that assumes only smoothness would need many more points than those sets hold. The positive-orthant solution is the maximum
likelihood estimate only for strictly positive densities. Streaming has its own limits: the burn-in dominates at small
$n$; above the true cut-off the held-out likelihood is nearly flat, so a cut-off that overshoots returns slowly; the
stopping rule was never met within 20 passes on the non-smooth truths (\NsConv\ of \NsRuns\ runs), so a fixed pass
budget is the practical rule; and the streams here are stationary.

\paragraph{Future work.} The theory needs only a compact frequency support for $\sqrt f$, symmetric for a real root:
a union of separate bands learned from the data would fit densities whose spectrum is sparse rather than low-pass.
Non-stationary streams need a forgetting rule, under which the counts change after the first epoch and the convergence
argument no longer applies.

\acks{
The authors received no specific funding for this work. Competing interests: R.A. is employed by Meta Platforms, Inc. A.A. is employed by Cohesity, Inc., and was previously employed by Rubrik, Inc. This work was carried out independently, outside the authors' employment; their employers had no role in it. The authors declare no other competing interests. The code is publicly available at
\url{https://github.com/rahuljhu/band-limited-ml} (MIT license), archived as release v1.0 at
\url{https://doi.org/10.5281/zenodo.24063291}; the per-experiment outputs are available from the
authors on request.}

\appendix
\section{Additional results}
\label{app:more}

\begin{table}[h]
  \centering\small
  \caption{Fitted slopes of log relative ISE against $\log n$ over $10^3\le n\le10^5$, with 95\%
  bootstrap intervals. For T3 the BLML and sinc-KDE rows use $f_c=2$ and $4$.}
  \label{tab:slopes}
  \fitwidth{%
\begin{tabular}{lcccccc}
\toprule
Estimator & $d=1$ & $d=2$ & $d=3$ & $d=4$ & $d=6$ & $d=8$ \\
\midrule
\multicolumn{7}{l}{\textit{separable (T1)}} \\
BLML, $f_c=f_c^{true}$ & $-1.05$\,{\scriptsize[-1.19,-0.89]} & $-0.94$\,{\scriptsize[-1.03,-0.85]} & $-1.00$\,{\scriptsize[-1.05,-0.94]} & $-1.02$\,{\scriptsize[-1.05,-0.98]} & $-0.96$\,{\scriptsize[-0.99,-0.93]} & $-0.89$\,{\scriptsize[-0.91,-0.88]} \\
BLML, $f_c=2f_c^{true}$ & $-1.00$\,{\scriptsize[-1.12,-0.86]} & $-0.97$\,{\scriptsize[-1.03,-0.90]} & $-1.01$\,{\scriptsize[-1.05,-0.97]} & $-0.98$\,{\scriptsize[-1.00,-0.97]} & $-0.78$\,{\scriptsize[-0.79,-0.78]} & $-0.49$\,{\scriptsize[-0.50,-0.48]} \\
KDE, oracle $h$ & $-0.67$\,{\scriptsize[-0.75,-0.59]} & $-0.66$\,{\scriptsize[-0.70,-0.61]} & $-0.58$\,{\scriptsize[-0.61,-0.54]} & $-0.49$\,{\scriptsize[-0.51,-0.47]} & $-0.37$\,{\scriptsize[-0.38,-0.36]} & $-0.28$\,{\scriptsize[-0.28,-0.27]} \\
KDE, LSCV $h$ & $-0.69$\,{\scriptsize[-0.77,-0.61]} & $-0.66$\,{\scriptsize[-0.70,-0.61]} & $-0.59$\,{\scriptsize[-0.61,-0.56]} & $-0.48$\,{\scriptsize[-0.50,-0.46]} & $-0.37$\,{\scriptsize[-0.38,-0.36]} & $-0.25$\,{\scriptsize[-0.28,-0.23]} \\
KDE, Scott $h$ & $-0.69$\,{\scriptsize[-0.77,-0.61]} & $-0.66$\,{\scriptsize[-0.71,-0.61]} & $-0.58$\,{\scriptsize[-0.61,-0.55]} & $-0.48$\,{\scriptsize[-0.50,-0.45]} & $-0.34$\,{\scriptsize[-0.35,-0.32]} & $-0.25$\,{\scriptsize[-0.26,-0.25]} \\
6th-order KDE, oracle $h$ & $-0.75$\,{\scriptsize[-0.86,-0.61]} & $-0.80$\,{\scriptsize[-0.86,-0.73]} & $-0.77$\,{\scriptsize[-0.81,-0.72]} & $-0.70$\,{\scriptsize[-0.72,-0.67]} & $-0.58$\,{\scriptsize[-0.60,-0.56]} & $-0.40$\,{\scriptsize[-0.42,-0.38]} \\
sinc KDE (kernel band = band of $f$) & $-0.97$\,{\scriptsize[-1.09,-0.84]} & $-0.97$\,{\scriptsize[-1.03,-0.91]} & $-1.01$\,{\scriptsize[-1.05,-0.98]} & $-1.00$\,{\scriptsize[-1.01,-0.98]} & $-1.00$\,{\scriptsize[-1.01,-0.99]} & $-1.00$\,{\scriptsize[-1.01,-0.99]} \\
Adaptive KDE & $-0.72$\,{\scriptsize[-0.79,-0.65]} & $-0.61$\,{\scriptsize[-0.64,-0.57]} & $-0.48$\,{\scriptsize[-0.50,-0.45]} & $-0.35$\,{\scriptsize[-0.38,-0.31]} & $-0.03$\,{\scriptsize[-0.06,0.01]} & $0.15$\,{\scriptsize[0.13,0.17]} \\
$k$-NN & $-0.51$\,{\scriptsize[-0.54,-0.48]} & $-0.52$\,{\scriptsize[-0.54,-0.50]} & $-0.53$\,{\scriptsize[-0.54,-0.51]} & $-0.46$\,{\scriptsize[-0.47,-0.44]} & $-0.30$\,{\scriptsize[-0.32,-0.28]} & $-0.21$\,{\scriptsize[-0.22,-0.20]} \\
\multicolumn{7}{l}{\textit{non-separable (T2)}} \\
BLML, $f_c=f_c^{true}$ & $-0.91$\,{\scriptsize[-0.98,-0.85]} & $-1.05$\,{\scriptsize[-1.09,-1.00]} & $-1.01$\,{\scriptsize[-1.04,-0.99]} & $-0.98$\,{\scriptsize[-0.99,-0.96]} & $-0.88$\,{\scriptsize[-0.89,-0.87]} & $-0.72$\,{\scriptsize[-0.72,-0.71]} \\
BLML, $f_c=2f_c^{true}$ & $-0.97$\,{\scriptsize[-1.02,-0.92]} & $-1.02$\,{\scriptsize[-1.04,-0.99]} & $-0.98$\,{\scriptsize[-0.99,-0.97]} & $-0.91$\,{\scriptsize[-0.92,-0.90]} & $-0.58$\,{\scriptsize[-0.59,-0.58]} & $-0.72$\,{\scriptsize[-0.75,-0.67]} \\
KDE, oracle $h$ & $-0.77$\,{\scriptsize[-0.81,-0.72]} & $-0.65$\,{\scriptsize[-0.69,-0.62]} & $-0.55$\,{\scriptsize[-0.57,-0.53]} & $-0.45$\,{\scriptsize[-0.47,-0.43]} & $-0.32$\,{\scriptsize[-0.33,-0.32]} & $-0.23$\,{\scriptsize[-0.23,-0.23]} \\
KDE, LSCV $h$ & $-0.75$\,{\scriptsize[-0.80,-0.71]} & $-0.66$\,{\scriptsize[-0.69,-0.62]} & $-0.55$\,{\scriptsize[-0.57,-0.53]} & $-0.45$\,{\scriptsize[-0.47,-0.43]} & $-0.32$\,{\scriptsize[-0.33,-0.32]} & $-0.23$\,{\scriptsize[-0.23,-0.23]} \\
KDE, Scott $h$ & $-0.60$\,{\scriptsize[-0.62,-0.58]} & $-0.38$\,{\scriptsize[-0.38,-0.38]} & $-0.23$\,{\scriptsize[-0.23,-0.23]} & $-0.13$\,{\scriptsize[-0.13,-0.13]} & $-0.03$\,{\scriptsize[-0.03,-0.03]} & $-0.01$\,{\scriptsize[-0.01,-0.01]} \\
6th-order KDE, oracle $h$ & $-0.84$\,{\scriptsize[-0.89,-0.79]} & $-0.83$\,{\scriptsize[-0.87,-0.78]} & $-0.74$\,{\scriptsize[-0.76,-0.72]} & $-0.67$\,{\scriptsize[-0.69,-0.64]} & $-0.49$\,{\scriptsize[-0.50,-0.48]} & $-0.36$\,{\scriptsize[-0.36,-0.35]} \\
sinc KDE (kernel band = band of $f$) & $-0.97$\,{\scriptsize[-1.02,-0.92]} & $-1.01$\,{\scriptsize[-1.03,-0.98]} & $-1.00$\,{\scriptsize[-1.01,-0.99]} & $-1.01$\,{\scriptsize[-1.02,-1.00]} & $-1.00$\,{\scriptsize[-1.00,-1.00]} & $-0.99$\,{\scriptsize[-1.00,-0.98]} \\
Adaptive KDE & $-0.75$\,{\scriptsize[-0.78,-0.71]} & $-0.62$\,{\scriptsize[-0.65,-0.58]} & $-0.50$\,{\scriptsize[-0.52,-0.48]} & $-0.38$\,{\scriptsize[-0.39,-0.36]} & $-0.03$\,{\scriptsize[-0.04,-0.02]} & $0.24$\,{\scriptsize[0.22,0.27]} \\
$k$-NN & $-0.54$\,{\scriptsize[-0.56,-0.51]} & $-0.58$\,{\scriptsize[-0.60,-0.56]} & $-0.51$\,{\scriptsize[-0.53,-0.50]} & $-0.42$\,{\scriptsize[-0.44,-0.40]} & $-0.30$\,{\scriptsize[-0.31,-0.29]} & $-0.21$\,{\scriptsize[-0.21,-0.20]} \\
\multicolumn{7}{l}{\textit{correlated normal (T3)}} \\
BLML, $f_c=2$ & $-1.02$\,{\scriptsize[-1.09,-0.94]} & $-1.02$\,{\scriptsize[-1.05,-0.99]} & $-1.00$\,{\scriptsize[-1.01,-0.98]} & $-0.93$\,{\scriptsize[-0.93,-0.92]} & $-0.59$\,{\scriptsize[-0.60,-0.57]} & $-0.74$\,{\scriptsize[-0.76,-0.72]} \\
BLML, $f_c=4$ & $-1.03$\,{\scriptsize[-1.11,-0.95]} & $-1.00$\,{\scriptsize[-1.03,-0.98]} & $-0.93$\,{\scriptsize[-0.94,-0.93]} & $-0.70$\,{\scriptsize[-0.71,-0.70]} & $-1.00$\,{\scriptsize[-1.09,-0.90]} & $-0.75$\,{\scriptsize[-0.93,-0.46]} \\
KDE, oracle $h$ & $-0.78$\,{\scriptsize[-0.84,-0.71]} & $-0.65$\,{\scriptsize[-0.68,-0.61]} & $-0.55$\,{\scriptsize[-0.58,-0.52]} & $-0.45$\,{\scriptsize[-0.47,-0.43]} & $-0.33$\,{\scriptsize[-0.34,-0.32]} & $-0.24$\,{\scriptsize[-0.25,-0.23]} \\
KDE, LSCV $h$ & $-0.86$\,{\scriptsize[-0.95,-0.75]} & $-0.66$\,{\scriptsize[-0.70,-0.62]} & $-0.55$\,{\scriptsize[-0.58,-0.52]} & $-0.46$\,{\scriptsize[-0.48,-0.44]} & $-0.33$\,{\scriptsize[-0.34,-0.32]} & $-0.24$\,{\scriptsize[-0.25,-0.23]} \\
KDE, Scott $h$ & $-0.77$\,{\scriptsize[-0.84,-0.71]} & $-0.64$\,{\scriptsize[-0.67,-0.60]} & $-0.54$\,{\scriptsize[-0.56,-0.51]} & $-0.43$\,{\scriptsize[-0.45,-0.41]} & $-0.30$\,{\scriptsize[-0.31,-0.29]} & $-0.20$\,{\scriptsize[-0.21,-0.20]} \\
6th-order KDE, oracle $h$ & $-0.89$\,{\scriptsize[-1.02,-0.78]} & $-0.81$\,{\scriptsize[-0.85,-0.77]} & $-0.76$\,{\scriptsize[-0.80,-0.72]} & $-0.68$\,{\scriptsize[-0.71,-0.65]} & $-0.51$\,{\scriptsize[-0.53,-0.49]} & $-0.37$\,{\scriptsize[-0.38,-0.36]} \\
sinc KDE, $f_c=4$ & $-1.02$\,{\scriptsize[-1.11,-0.94]} & $-1.00$\,{\scriptsize[-1.02,-0.97]} & $-1.00$\,{\scriptsize[-1.00,-0.99]} & $-1.00$\,{\scriptsize[-1.00,-1.00]} & $-1.03$\,{\scriptsize[-1.07,-1.00]} & $-0.93$\,{\scriptsize[-0.98,-0.89]} \\
Adaptive KDE & $-0.72$\,{\scriptsize[-0.78,-0.66]} & $-0.63$\,{\scriptsize[-0.66,-0.59]} & $-0.51$\,{\scriptsize[-0.53,-0.49]} & $-0.40$\,{\scriptsize[-0.41,-0.39]} & $-0.06$\,{\scriptsize[-0.07,-0.05]} & $0.26$\,{\scriptsize[0.23,0.29]} \\
$k$-NN & $-0.53$\,{\scriptsize[-0.56,-0.51]} & $-0.54$\,{\scriptsize[-0.56,-0.51]} & $-0.52$\,{\scriptsize[-0.53,-0.50]} & $-0.44$\,{\scriptsize[-0.46,-0.43]} & $-0.30$\,{\scriptsize[-0.32,-0.29]} & $-0.21$\,{\scriptsize[-0.22,-0.20]} \\
\bottomrule
\end{tabular}%
}
\end{table}

\begin{figure}[h]
  \centering
  \includegraphics[width=\textwidth]{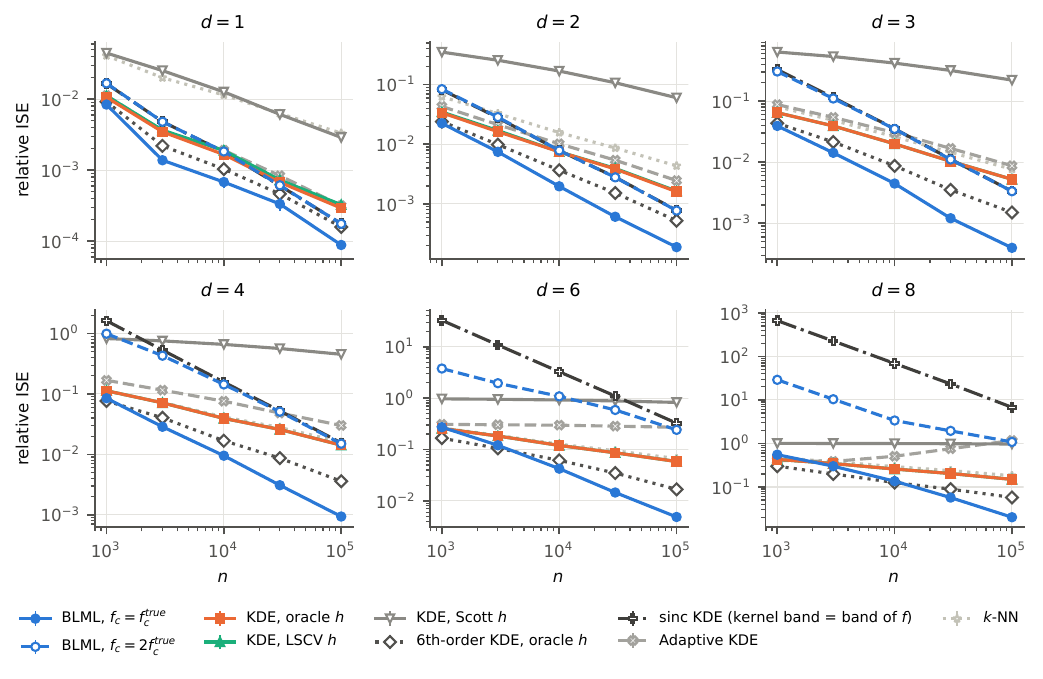}
  \caption{Relative ISE against $n$ for the non-separable truth T2.}
  \label{fig:rate-t2}
\end{figure}

\begin{figure}[h]
  \centering
  \includegraphics[width=0.6\textwidth]{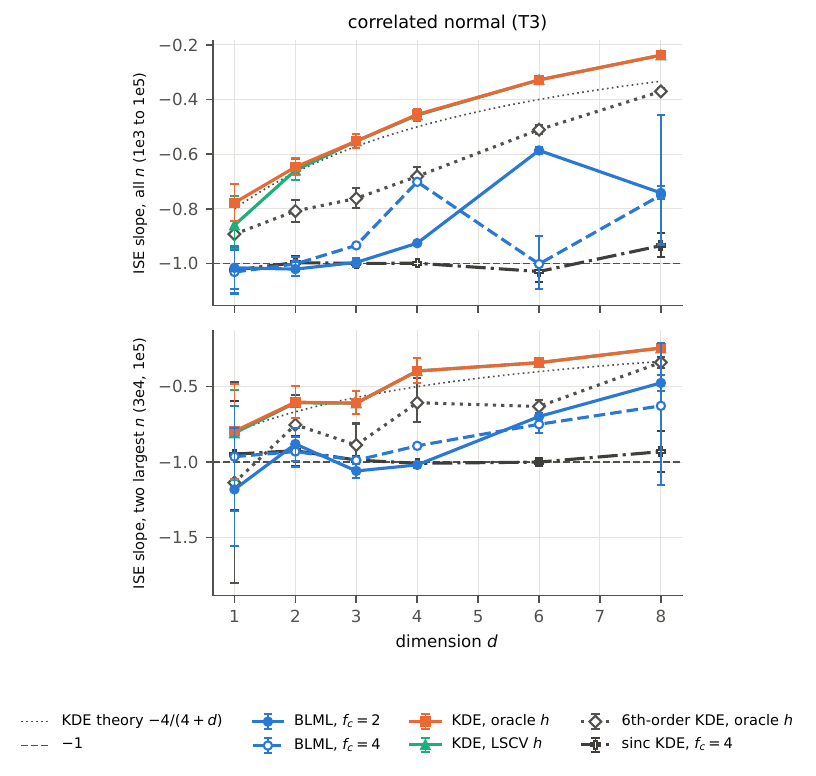}
  \caption{Fitted slopes for the correlated normal T3 (not band-limited).}
  \label{fig:slopes-t3}
\end{figure}

\begin{figure}[h]
  \centering
  \includegraphics[width=\textwidth]{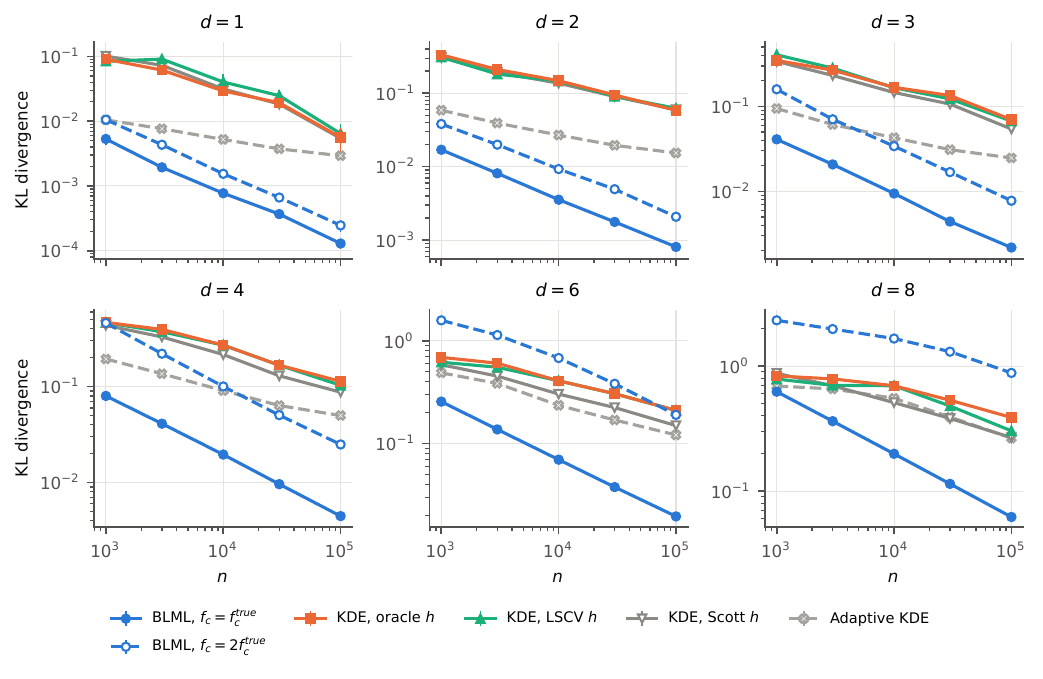}
  \caption{Kullback--Leibler divergence against $n$, T1.}
  \label{fig:kl-t1}
\end{figure}

\begin{figure}[ht]
  \centering
  \includegraphics[width=\textwidth]{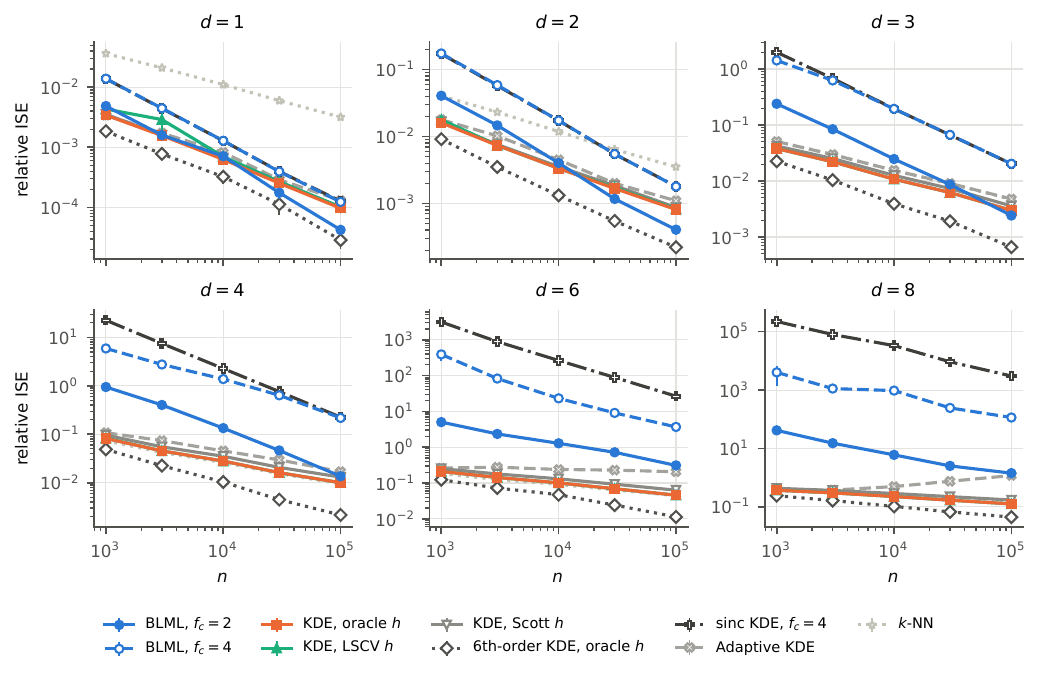}
  \caption{Relative ISE against $n$ for the correlated normal T3 ($f_c=2$ and $4$ for BLML and
  sinc KDE).}
  \label{fig:rate-t3}
\end{figure}

\begin{figure}[ht]
  \centering
  \includegraphics[width=\textwidth]{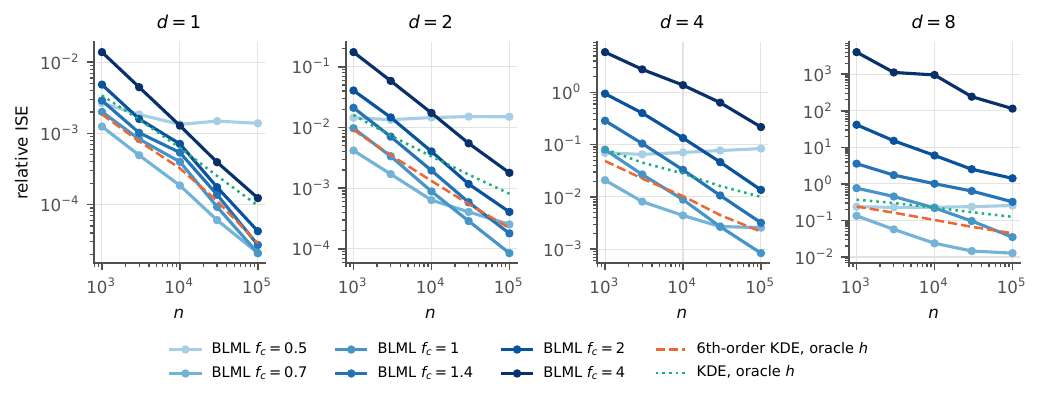}
  \caption{Correlated normal T3: relative ISE against $n$ for BLML at six cutoffs (light to dark:
  low to high) and the oracle-bandwidth kernel estimators. Low cutoffs win at small $n$ and level
  off at a bias floor; high cutoffs decay as $n^{-1}$ with a large constant.}
  \label{fig:t3-cutoffs}
\end{figure}

\subsection{Solver: convergence, cost and precision}
\label{app:solver}

\begin{figure}[ht]
  \centering
  \includegraphics[width=0.95\textwidth]{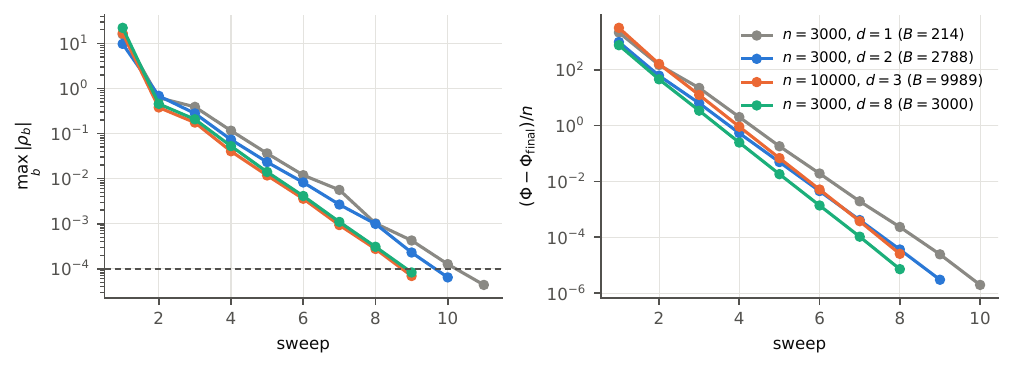}
  \caption{Convergence: constraint residual $\max_b|\rho_b|$ (left; dashed: tolerance) and
  objective gap (right) per sweep.}
  \label{fig:conv}
\end{figure}

\paragraph{Convergence.}
Every fit we report converged to $\max_b|\rho_b|<10^{-4}$ with a monotonically decreasing
objective. The residual and the objective gap both fall geometrically, by roughly a factor
of three per sweep (Fig.~\ref{fig:conv}), and the solver needs 9--11 sweeps regardless of
$n$ or $d$. The number of sweeps does not grow with dimension; it is 6--10 for $d=2$ to $16$ at
$n=3000$ and 8--9 at $n=30{,}000$. Without the descent safeguard the
one-dimensional fit oscillated and needed \UnsafeSweepsDOne\ sweeps; with it, 11. The
safeguard costs nothing on its normal path, because the objective is evaluated once per
sweep anyway.

\begin{figure}[ht]
  \centering
  \includegraphics[width=0.5\textwidth]{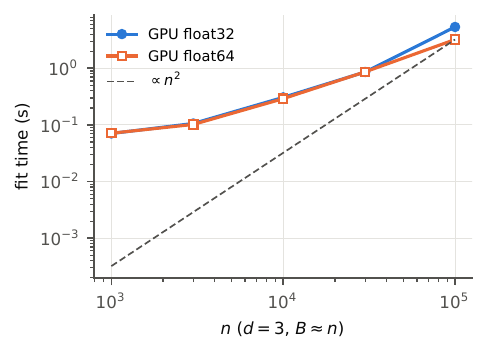}
  \caption{GPU fit time against $n$ ($d=3$, where $B\approx n$), float32 and float64.}
  \label{fig:time}
\end{figure}

\paragraph{Precision.}
float32 and float64 run at comparable speed on the A100 (Fig.~\ref{fig:time}), because the
solver is limited by memory traffic and launches rather than arithmetic. Agreement
with a tightly converged float64 reference is set by the solver tolerance, not by the
precision: at tolerance $10^{-4}$ both precisions give the same maximum relative deviation
($3.8\times10^{-5}$ where $\hat f$ exceeds $10^{-3}$ of its maximum); at $10^{-7}$ float64 reaches
$2.2\times10^{-8}$ (Table~\ref{tab:prec}). Larger relative deviations occur only where $\hat f$
is nearly zero.

\begin{table}[ht]
  \centering\small
  \caption{Deviation from a float64 fit converged to tolerance $10^{-9}$ ($n=3000$, $d=3$,
  $2\times10^4$ test points).}
  \label{tab:prec}
  \fitwidth{%
\begin{tabular}{lrrrr}
\toprule
setting & max rel.\ diff & max rel.\ diff where $\hat f>10^{-3}\max\hat f$ & sweeps & converged \\
\midrule
float32, tol $=$ 0.0001 & 3.1e-04 & 3.8e-05 & 10 & yes \\
float32, tol $=$ 1e-06 & 8.7e-05 & 7.6e-07 & 15 & yes \\
float64, tol $=$ 0.0001 & 3.1e-04 & 3.8e-05 & 10 & yes \\
float64, tol $=$ 1e-07 & 9.1e-07 & 2.2e-08 & 16 & yes \\
\bottomrule
\end{tabular}%
}
\end{table}

\paragraph{Block size.}
Larger blocks amortize the per-block kernel launch: going from $B_s=32$ to $512$ cuts fit time
twelve-fold ($2.12$ to $0.17$\,s at $n=10^4$, $d=3$) with the same number of sweeps and the
same final objective to nine digits (Table~\ref{tab:block}).

\begin{table}[ht]
  \centering\small
  \caption{CPU (Numba, single thread) versus GPU (A100), same algorithm and tolerance.}
  \label{tab:speed}
  \begin{tabular}{lrrrrrr}
\toprule
Workload & $B$ & \multicolumn{2}{c}{fit (s)} & \multicolumn{2}{c}{predict (s)} & speedup \\
 & & CPU & GPU & CPU & GPU & fit / predict \\
\midrule
synthetic d=3 n=3000 & 2,999 & 1.6 & 0.13 & 1.7 & 0.007 & 12$\times$ / 237$\times$ \\
synthetic d=3 n=10000 & 9,991 & 12.1 & 0.38 & 5.3 & 0.021 & 32$\times$ / 256$\times$ \\
synthetic d=3 n=30000 & 29,963 & 160.9 & 0.87 & 15.9 & 0.059 & 185$\times$ / 269$\times$ \\
CAISO p(price|hour) n=10001 & 9,808 & 19.8 & 0.29 & 5.4 & 0.024 & 69$\times$ / 221$\times$ \\
\bottomrule
\end{tabular}
\end{table}

\begin{table}[ht]
  \centering\small
  \caption{Peak GPU memory of a BLML fit against the size of a dense float32 Gram matrix.}
  \label{tab:mem}
  \fitwidth{%
\begin{tabular}{rrrrr}
\toprule
$d$ & $n$ & $B$ & BLML GPU peak & dense Gram (float32) \\
\midrule
3 & 1,000 & 1,000 & 5 MB & 0.00 GB \\
3 & 3,000 & 2,998 & 15 MB & 0.04 GB \\
3 & 10,000 & 9,992 & 52 MB & 0.40 GB \\
3 & 30,000 & 29,966 & 155 MB & 3.59 GB \\
3 & 100,000 & 99,894 & 516 MB & 39.92 GB \\
8 & 1,000 & 1,000 & 5 MB & 0.00 GB \\
8 & 3,000 & 3,000 & 16 MB & 0.04 GB \\
8 & 10,000 & 10,000 & 52 MB & 0.40 GB \\
8 & 30,000 & 30,000 & 156 MB & 3.60 GB \\
8 & 100,000 & 100,000 & 521 MB & 40.00 GB \\
\bottomrule
\end{tabular}%
}
\end{table}

\begin{table}[ht]
  \centering\small
  \begin{minipage}{0.42\textwidth}\centering
  \caption{Block size ($n=10^4$, $d=3$).}
  \label{tab:block}
  \fitwidth{%
\begin{tabular}{rrrc}
\toprule
block size $B_s$ & sweeps & fit time (s) & monotone $\Phi$ \\
\midrule
32 & 9 & 2.12 & yes \\
64 & 9 & 1.10 & yes \\
128 & 9 & 0.56 & yes \\
256 & 9 & 0.29 & yes \\
512 & 10 & 0.17 & yes \\
\bottomrule
\end{tabular}%
}
  \end{minipage}\hfill
  \begin{minipage}{0.55\textwidth}\centering
  \caption{Lattice refinement: relative ISE (mean $\pm$ s.d., 5 replicates).}
  \label{tab:bin}
  \fitwidth{%
\begin{tabular}{rrcc}
\toprule
$d$ & $n$ & rel.\ ISE, $h_k=0.1n^{-1/4}/f_c$ & rel.\ ISE, $h_k=0.025n^{-1/4}/f_c$ \\
\midrule
1 & 1,000 & 8.95e-04 $\pm$ 5.7e-04 & 8.92e-04 $\pm$ 5.6e-04 \\
1 & 10,000 & 4.08e-05 $\pm$ 1.7e-05 & 4.05e-05 $\pm$ 1.7e-05 \\
2 & 1,000 & 1.54e-03 $\pm$ 5.1e-04 & 1.54e-03 $\pm$ 5.1e-04 \\
2 & 10,000 & 1.59e-04 $\pm$ 2.3e-05 & 1.60e-04 $\pm$ 2.3e-05 \\
\bottomrule
\end{tabular}%
}
  \end{minipage}
\end{table}

\subsection{Kullback--Leibler divergence}
\label{app:kl}

\begin{figure}[ht]
  \centering
  \includegraphics[width=\textwidth]{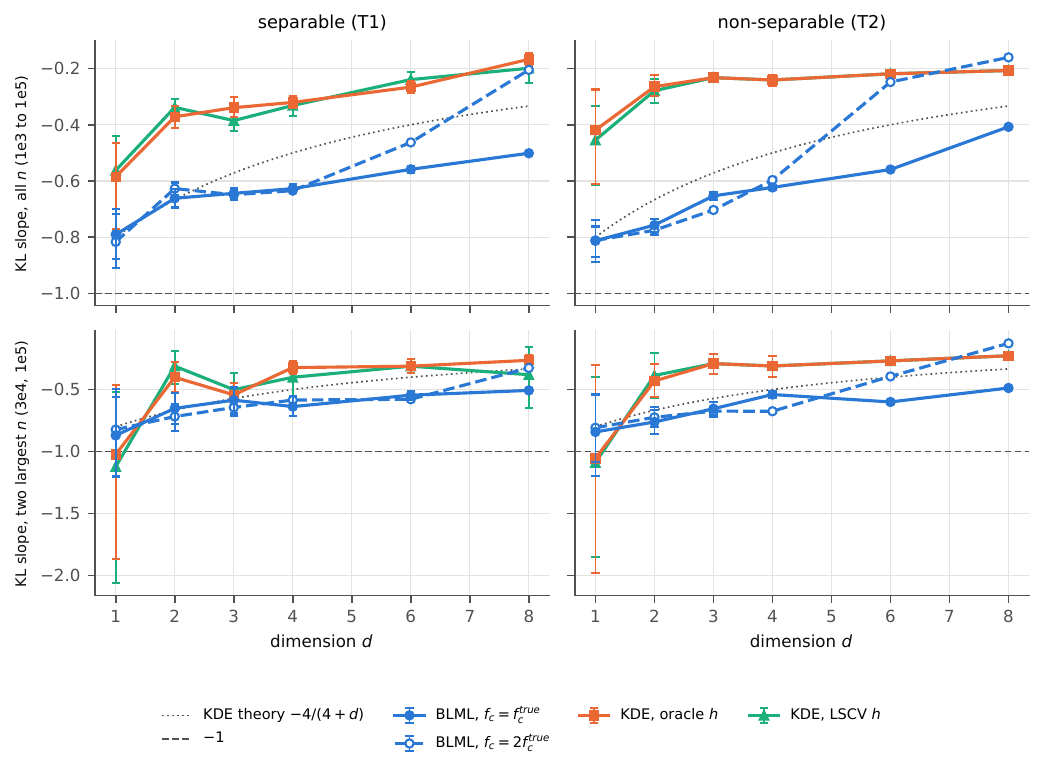}
  \caption{Fitted slope of log KL divergence against $\log n$ (top: all $n$; bottom: the two largest
  $n$), with 95\% bootstrap intervals, for estimators that return a density. Dashed: $-1$.}
  \label{fig:slopes-kl}
\end{figure}

KL weights errors by $1/f$, so it is dominated by the sparse tails, and every estimator's
KL decays more slowly than its ISE (Fig.~\ref{fig:slopes-kl}). BLML at $f_c^{\text{true}}$ has
the lowest KL of all estimators that return a valid density at every $n$ and $d$ for T1; at
$n=10^5$ it is $6$--$72\times$ below the oracle Gaussian KDE. On T2 the adaptive KDE has the
lower KL at small $n$ for $d\ge3$, and at every $n$ for $d=8$ ($0.26$ against $0.28$ at $n=10^5$). KL
is undefined for the signed estimators, which are negative at some sample points, and
meaningless for $k$-NN, which is not normalized, so neither appears in the KL figures.
BLML is non-negative by construction and integrates to one up to the solver tolerance
(Proposition~\ref{prop:convex}).

\section{The rate as a conjecture}
\label{app:conjecture}

Section~\ref{sec:rate} argues for the rate heuristically and Section~\ref{sec:experiments} measures it. Here we state
precisely what we believe holds and what a proof would need, so that the claim can be checked and, we hope, proved.

\paragraph{Setting.} Let $x_1,\dots,x_n$ be i.i.d.\ from $f=g^2$, where $g\in\mathcal V$ is band-limited to the box
$\Omega$ of Section~\ref{sec:background}, has no zeros, and $\int f(x)\,\|x\|^{2+\delta}dx<\infty$ for some
$\delta>0$. Let $\hat f_n$ be the positive-orthant BLML estimate at a fixed cut-off $f_c$ whose box contains $\Omega$.

\begin{conjecture}\label{conj:rate}
For every fixed $d$ there is a constant $C(f,f_c)<\infty$ with
$\mathbb E\int(\hat f_n-f)^2 = C(f,f_c)\,n^{-1}+o(n^{-1})$, and
the relative error $C(f,f_c)/\int f^2$ grows in proportion to $p_{\text{eff}}=\prod_k f_{c,k}R_k$, with
$R_k=1/\int f_k^2$ for a product density.
\end{conjecture}

The measurements of Section~\ref{sec:exp-rate} are consistent with both parts: slopes \SlopeSummary, and a
constant growing as about $1.85^d$ on T1, where $\prod_kf_{c,k}R_k$ grows by a fixed factor per dimension.

\paragraph{What a proof needs.} We see three steps; we have not carried them out, and each may hide difficulties.
(i)~\emph{The right root.} The positive-orthant solution must be the global maximiser with probability tending to
one; \citet{agarwal2017blml} show this asymptotically for $f>0$, which is why $g$ is assumed to have no zeros.
(ii)~\emph{A finite effective dimension.} Restricted to a box carrying all but $\varepsilon$ of the mass, the
time- and band-limiting operator has about $p_{\text{eff}}$ eigenvalues near one and a tail that decays faster than
geometrically \citep{landau1980eigenvalue}. A sieve argument would project $g$ and $\hat g$ on the leading
$p_{\text{eff}}+m$ eigenfunctions, bound the remainder by the tail eigenvalues and the mass outside the box (here the
moment condition enters), and choose $m$ and $\varepsilon$ growing slowly with $n$ so that the remainder is
$o(n^{-1})$. (iii)~\emph{Parametric asymptotics on the sieve.} On the leading space the estimate is a constrained
maximum likelihood estimate of a finite-dimensional $g$ with $\|g\|_2=1$, for which the usual expansion gives an
$L^2$ error of order (dimension)$/n$ for $\hat g$ and, through $\hat f-f=(\hat g-g)(\hat g+g)$, for $\hat f$. The hard
part is uniformity: the sieve dimension grows with $n$, so the expansion must hold with constants that do not.

\section{Conditional densities}
\label{app:prices}

\paragraph{Conditional density.} A conditional density $p(y\mid x)$ is estimated as the ratio of two BLML fits with the same
cutoffs, the joint fit $\hat f_J(y,x)$ and a fit $\hat f_M(x)$ to the covariates alone, $\hat f(y\mid x)=\hat
f_J(y,x)/\hat f_M(x)$, as the Nadaraya--Watson conditional KDE is a joint KDE divided by a marginal KDE. The two fits
share the lattice and the kernel blocks: the solver forms each block once and uses its product over all dimensions
for the joint and over the covariate dimensions for the marginal. The ratio integrates over $y$ to $\tilde
f_M(x)/\hat f_M(x)$, where $\tilde f_M=\int\hat f_J\,dy$ is the marginal implied by the joint. Writing $S$ for the
expected log score of the ratio and $S_{\mathrm{norm}}$ for that of the exactly normalised conditional $\hat
f_J/\tilde f_M$,
\begin{equation}
  S-S_{\mathrm{norm}}=\mathrm{KL}(p_M\,\|\,\hat f_M)-\mathrm{KL}(p_M\,\|\,\tilde f_M),
\end{equation}
so whenever the direct marginal fit is the more accurate of the two estimates of the covariate density, the reported
score is below that of the normalised conditional and is conservative.

\section{Flows of other sizes}
\label{app:flow-sizes}

\begin{figure}[t]
  \centering
  \includegraphics[width=0.62\textwidth]{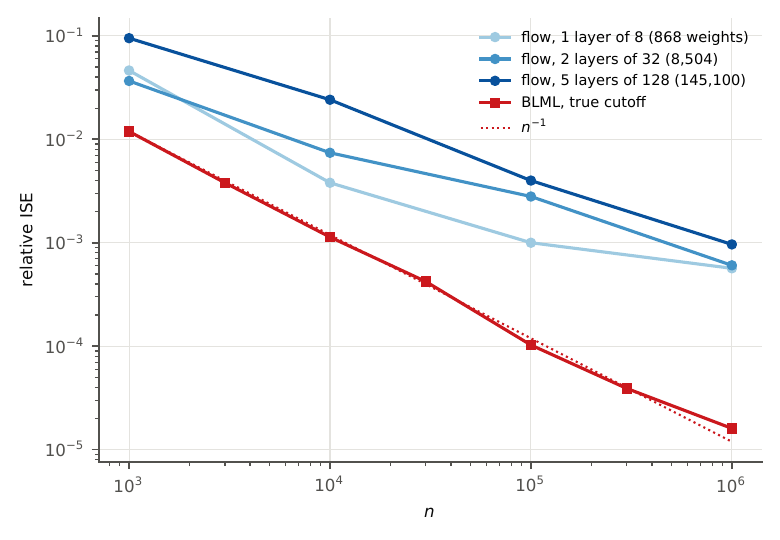}
  \caption{Fixed architectures against sample size on the band-limited truth T1 at $d=4$: neural spline flows of three
  sizes (weights in parentheses) and BLML at the true cutoff, with $n^{-1}$ for reference. One replicate at
  $n=10^6$, where the smallest flow reached its cap of 150 passes; two below.}
  \label{fig:flow-sizes}
\end{figure}

Figure~\ref{fig:flow-sizes} varies the flow's size. No single size is right across $n$: the largest flow, with
145{,}100 weights, is the least accurate of the three up to $n=10^5$, its variance not yet paid for by the data, while
the smallest improves only 1.8-fold from $n=10^5$ to $10^6$ (that run reached its pass cap, so part of the slowdown
may be training rather than capacity). BLML follows $n^{-1}$ throughout and is 35 to 60 times more accurate than any
of them at $n=10^6$.

\section{The one-shot search}
\label{app:oneshot}

\paragraph{Stage 1: shared search on a volume grid.} Starting from $f=0.2\cdot\mathbf 1$, all cut-offs
are doubled together until $\mathrm{VLL}$ fails to improve for three consecutive steps (the patience
guards against validation noise). Doubling every cut-off multiplies $\prod_jf_{c,j}$, which sets BLML's
variance through the effective number of parameters, by $2^d$, a step far too coarse in high dimension.
The bracket is therefore refined on the grid $f\cdot2^{j/d}$, $|j|<d$, around the best doubling, on which
$\prod_jf_{c,j}$ changes by a factor of two per step at every $d$, walked outward from the best doubling
and stopped at the first step that does not improve.

\paragraph{Stage 2: per-dimension refinement.} From the best shared vector, coordinate sweeps multiply
each $f_{c,j}$ by $\sqrt2$ while that improves $\mathrm{VLL}$, or else divide by $\sqrt2$ while that does,
until a full sweep accepts no move. A step of $\sqrt2$ suffices because the volume grid has already
placed the shared optimum to within a factor $2^{1/d}$.

\paragraph{Periodic dimensions move in whole harmonics.} A periodic kernel with period $P$ and cut-off
$f_c$ has $N=\lfloor f_cP/2\rfloor$ harmonics, so $\mathrm{VLL}$ is piecewise constant in $f_c$: a step
that leaves $N$ unchanged sees no change and stops a local search. Any proposed move that leaves $N$
unchanged is extended to the next harmonic count in its direction.

\section{Experimental details}
\label{app:setup}

\paragraph{Sampling.} $\mathrm{sinc}^2$ is drawn by rejection from a Cauchy proposal (acceptance $1/2$), $\mathrm{sinc}^4$
and $s^2$ by further rejection, and $h^2$ by rejection from the mixture $\sum_ja_jK_j^2$, which dominates
$h^2/\sum_ja_j$ by Cauchy--Schwarz. The samplers agree with grid quadrature ($\int f^2$ within 0.1\%). For T2,
$h(x)=\sum_{j=1}^4a_j\prod_ks(0.2(x_k-\mu_{jk}))$ with $s(t)=\mathrm{sinc}^2(t)+\mathrm{sinc}^2(t+0.5)$, positive
weights and centres on the diagonal; $s>0$ because its two terms have disjoint zeros.

\paragraph{Error estimates.} For $d\le2$ the ISE is a Riemann sum. For BLML it is exact up to tail truncation:
$(\hat f-f)^2$ is band-limited to $2f_c$, and a uniform sum integrates a function band-limited to $W$ exactly when the
spacing is below $1/W$. For the kernel estimators we use spacings of $0.05$ ($d=1$) and $0.25$ ($d=2$), well below
their bandwidths, over ranges of $\pm400$ and $\pm80$. For $d>2$ we importance-sample $2\times10^5$ points from
$q=\tfrac12f+\tfrac12\prod_k\mathrm{Cauchy}(0,6)$, which keeps $(\hat f-f)^2/q$ bounded where $f$ has mass and in its
$|x|^{-4}$ tails. Hellinger and Kullback--Leibler divergences use $10^5$ draws from $f$.

\paragraph{Signed estimators.} At points drawn from T1 the sinc KDE is negative at $0.02$--$0.3\%$ of them in $d=1$
and $16$--$47\%$ in $d=8$; the sixth-order KDE at up to $4.6\%$ (T1) and $7.5\%$ (T3).
\section{Cost details}
\label{sec:costdetail}

\paragraph{Fixed parameters.} When no kernel parameter is learned (cut-offs frozen, no trainable map
or period), a batch adds its counts, computes the touched rows and takes the coordinate steps; the
prequential score, its gradient, the leave-batch-out removal and the $Q$ ratio are skipped. The rows
$(Se)_I$ need no derivatives, so they are computed as one product of a $|I|\times B$ kernel block, built
from the table entries in chunks, with the coefficients (float64 sums on the GPU), three times faster than
the pair-by-pair accumulation that also returns derivatives; for a conditional model the block of the
marginal is built by masking the dependent columns. An epoch then does the work of one sweep of the exact
solver (Section~\ref{sec:fixed}).

\paragraph{Precision on a GPU.} The pair kernels run in float32 arithmetic with float64 sums when the
estimator's dtype is float32 (GPUs with slow float64, such as the T4), and in float64 otherwise; the kernel
tables are built on the device and cached until a parameter moves.

\paragraph{Tables.} A table row has one entry per lattice offset, $2J+1$ for a sinc column where $J$ is the
bin span in lattice steps, independent of $B$. Heavy-tailed data make $J$ large (millions at $10^5$
points), and building rows on the host and uploading them took about half the GPU time
(15.7 of 34.5\,s at $4\cdot10^4$ points); building them on the device removed it. Radial (ball) tables are
indexed by squared lattice distance, whose range grows with the square of the span; we tabulate them on a
grid in $r$ with linear interpolation instead (relative error $<10^{-5}$), so their size is linear in the
span, and build them on the device from an elementwise Bessel routine. A column with few distinct bin
values $v$ ($|v|^2<2J+1$: hours, integer data) stores instead the $|v|^2$ pairs of its occupied
values, so a column's table never exceeds $\min(2J+1,|v|^2)$ entries. Fewer offsets than either actually
occur: on the benchmark of Section~\ref{sec:speed} a dense row holds $2\cdot10^6$--$1.6\cdot10^7$ entries
and only $1$--$5\cdot10^5$ distinct offsets are used. A table of those alone would need a search per lookup
in a loop of $10^9$--$10^{10}$ lookups per sweep, whereas a dense row is indexed directly by the offset; it
is rebuilt whenever a parameter moves, which took a few percent of a streaming run at $2\cdot10^4$ points.
\section{Ablations}
\label{sec:ablation}

\paragraph{Gradient corrections.} Table~\ref{tab:ablation} adds the gradient corrections one at a time, on the
benchmark of Section~\ref{sec:speed} at $2\cdot10^4$ points. On this
benchmark each correction helps in turn. With the coefficients held fixed, a step after every batch and no
smoothing, the gradient ends 0.04 nats per point behind one-shot selection, with the first cut-off more than twice too
high; the one-sweep response brings it to within 0.012, smoothing to within 0.007, and pooling over three batches
matches one-shot selection, at about a quarter of its time. These are single runs, and the steps after the first are
within the spread over seeds (about 0.015 nats per point). Once smoothing and pooling are in place, the one-sweep
response no longer matters: over five seeds, switching it on changes the held-out score by $+0.005$, $-0.004$,
$-0.001$, $+0.007$ and $-0.015$ nats per point (mean $-0.0015$) at the same cost. The default is therefore smoothing
and pooling without it.

\begin{table}[t]
\centering
\caption{Gradient variants for streaming BLML, each added to the one above.}
\label{tab:ablation}
\begin{tabular}{lcc}
\toprule
variant ($n=2\cdot10^4$, CPU, two passes) & time (s) & held-out log-lik. \\
\midrule
one-shot selection (reference) & 83.5 & $-2.4214$ \\
fixed-e gradient & 24.8 & $-2.4629$ \\
\quad + one-sweep response & 24.2 & $-2.4332$ \\
\quad + smoothing & 22.4 & $-2.4285$ \\
\quad + pooled over 3 batches & 21.7 & $-2.4209$ \\
default: smoothing and pooling, no one-sweep response & 16.6 & $-2.4257$ \\
\bottomrule
\end{tabular}
\end{table}

\paragraph{Revisit corrections.} On T1 at $d=4$, $n=2\cdot10^4$ (two replicates, four passes, CPU) we
remove the corrections of revisited batches one at a time. Without the leave-batch-out removal a revisited
batch scores the model that already contains it, the score rewards over-fitting, and the cut-offs creep up
epoch by epoch to $0.57$--$0.66$, with a final error 5--6 times the full method's ($4.3$ and $3.8\cdot10^{-3}$
against $6.7$ and $7.9\cdot10^{-4}$). Without the derivative of the normaliser the cut-offs wander (one
dimension to $0.64$). The correction of the tracked $Q$ toward its estimate changes nothing measurable here
and costs almost nothing; we keep all three.

\section{Learning a period}
\label{sec:period}

\begin{figure}[t]
  \centering
  \includegraphics[width=0.55\textwidth]{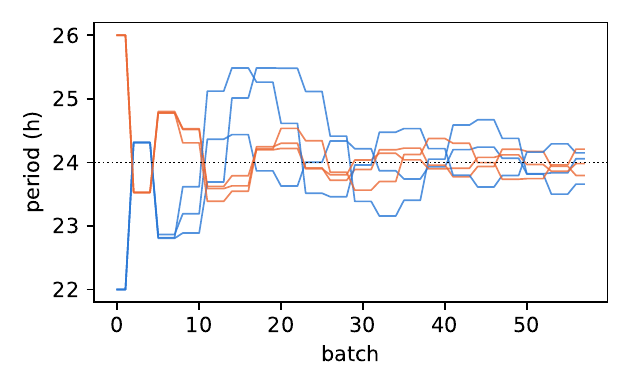}
  \caption{Learned period against batch, from 22\,h (blue) and 26\,h (orange), three seeds each; dotted:
  24\,h.}
  \label{fig:period}
\end{figure}

 On a ten-day record with a daily cycle, $z=\sin(2\pi t/24)+0.4\varepsilon$
with $t$ uniform on $[0,240)$\,h ($4\cdot10^4$ points, batches of 2000, three passes), we fit the
conditional density of $z$ given $t$, with $t$ a Dirichlet column of trainable period on the unfolded
lattice. Started at 22\,h and at 26\,h with three seeds each, the period ends at
\PeriodRange\,h (mean \PeriodMean; Fig.~\ref{fig:period}), in about \PeriodTime\,s per run on a CPU. The
conditional matters: in a joint fit the marginal of $t\bmod P$ carries $-\log P$, and a joint run walked to
8\,h (Section~\ref{sec:grad}).

\section{The same learning applied to a KDE: tables}
\label{app:kde}

\begin{table}[t]
  \centering\scriptsize\setlength{\tabcolsep}{3pt}
  \caption{One-dimensional ISE at $n=64{,}000$ (best in bold) and the least-squares slope of
  $\log$ ISE in $\log n$ over $n=2{,}000$--$64{,}000$. The KDE theory gives $-0.8$.}
  \label{tab:kde_ise}
  \begin{tabular}{ll rrrrr rrrrr}
\toprule
 & & \multicolumn{5}{c}{ISE at $n=64{,}000$} & \multicolumn{5}{c}{slope of $\log$ ISE in $\log n$} \\
\cmidrule(lr){3-7}\cmidrule(lr){8-12}
truth & seed & str.\ KDE & oracle & valid. & Scott & str.\ BLML & str.\ KDE & oracle & valid. & Scott & str.\ BLML \\
\midrule
T1 & 0 & 1.2e-3 & \textbf{2.2e-5} & 2.8e-5 & 2.4e-5 & 2.4e-5 & $+0.41$ & $-0.77$ & $-0.76$ & $-0.80$ & $-1.33$ \\
T1 & 1 & 6.7e-4 & 2.3e-5 & 1.1e-4 & 2.3e-5 & \textbf{3.5e-6} & $+0.36$ & $-0.77$ & $-0.78$ & $-0.77$ & $-1.77$ \\
\midrule
T2 & 0 & 4.3e-5 & 1.8e-5 & 3.6e-5 & 1.8e-4 & \textbf{9.3e-6} & $-1.08$ & $-0.85$ & $-0.99$ & $-0.62$ & $-1.19$ \\
T2 & 1 & 8.7e-5 & 2.0e-5 & 4.1e-5 & 1.9e-4 & \textbf{5.1e-6} & $-0.79$ & $-0.77$ & $-0.90$ & $-0.55$ & $-1.33$ \\
\midrule
T3 & 0 & 6.4e-5 & 5.8e-5 & 8.5e-5 & 5.8e-5 & \textbf{9.9e-6} & $-0.54$ & $-0.54$ & $-0.53$ & $-0.61$ & $-2.32$ \\
T3 & 1 & 7.7e-5 & 6.1e-5 & 1.6e-4 & 7.5e-5 & \textbf{2.5e-5} & $-0.39$ & $-0.35$ & $-0.17$ & $-0.38$ & $-2.19$ \\
\bottomrule
\end{tabular}
\end{table}

\begin{table}[t]
  \centering\scriptsize\setlength{\tabcolsep}{3pt}
  \caption{Excess test loss (nats, lower is better; best in bold) at $n=64{,}000$, and streaming
  KDE's bandwidth relative to the oracle's. Differences below about $10^{-3}$ are within the noise
  of the $5{,}000$-point test sample, which also makes some entries negative.}
  \label{tab:kde_kl}
  \begin{tabular}{lll rrrrr r}
\toprule
truth & $d$ & seed & streaming KDE & KDE, oracle $h$ & KDE, validated $h$ & KDE, Scott & streaming BLML & $h_{\rm str}/h_{\rm or}$ \\
\midrule
T1 & 1 & 0 & 0.01590 & 0.02723 & 0.15171 & 0.10722 & \textbf{0.00170} & 2.01 \\
T1 & 1 & 1 & 0.00509 & 0.00173 & 0.00173 & 0.00533 & \textbf{0.00113} & 1.66 \\
T2 & 1 & 0 & 0.00553 & 0.00415 & 0.00623 & 0.00468 & \textbf{0.00134} & 0.76 \\
T2 & 1 & 1 & 0.00015 & \textbf{-0.00011} & -0.00011 & 0.00139 & 0.00148 & 1.31 \\
T3 & 1 & 0 & 0.00007 & 0.00004 & 0.00013 & 0.00004 & \textbf{-0.00011} & 1.22 \\
T3 & 1 & 1 & -0.00009 & -0.00016 & -0.00013 & -0.00009 & \textbf{-0.00024} & 2.78 \\
\midrule
T1 & 2 & 0 & 0.03184 & 0.00300 & 0.00317 & 0.00428 & \textbf{0.00098} & 2.38 \\
T1 & 2 & 1 & 0.01988 & 0.00428 & 0.00782 & 0.00782 & \textbf{0.00197} & 1.89 \\
T2 & 2 & 0 & 0.00767 & 0.00578 & 0.00719 & 0.08219 & \textbf{0.00214} & 1.27 \\
T2 & 2 & 1 & 0.00551 & 0.00438 & 0.00438 & 0.06317 & \textbf{0.00411} & 1.25 \\
T3 & 2 & 0 & 0.00158 & 0.00159 & 0.00159 & 0.00159 & \textbf{0.00104} & 1.04 \\
T3 & 2 & 1 & 0.00185 & 0.00158 & 0.00158 & 0.00236 & \textbf{-0.00035} & 0.71 \\
\bottomrule
\end{tabular}
\end{table}
\section{A coarser re-binning tolerance}
\label{app:rebin}

The lattice need only stay far below the Nyquist spacing $1/(2f_c)$; $h=\beta n^{-1/4}/f_c$ with
$\beta=0.1$ leaves a wide margin. We therefore rebuilt it only when its target width moves by a factor 2
(the data grow 16-fold, or $f_c$ doubles) instead of $2^{1/4}$, on the benchmark of Section~\ref{sec:speed}
(CPU, two passes, burn-in on $5\cdot10^3$ points). Table~\ref{tab:rebin} compares it with the tolerance of
Section~\ref{sec:exp}, with and without the two sweeps after each rebuild. With the factor 2 no rebuild
occurs at these sizes (the data grow 4--8-fold after the burn-in), so the sweeps play no part. Its
held-out log-likelihood matches the $2^{1/4}$ tolerance with sweeps at $2\cdot10^4$ points (mean
$-2.4156$ against $-2.4196$ over three samples) and is $0.006$ lower on the one sample at $4\cdot10^4$,
within the spread over samples ($\approx0.015$); it is 7--18\% faster, the lattice at $4\cdot10^4$ holding
$35.4$ rather than $38.5$ thousand bins. Without sweeps the $2^{1/4}$ tolerance is $0.014$--$0.017$ behind:
the coefficients lag each rebuild and bias the cut-off gradient, which the coarser tolerance avoids by
rebuilding less often. The software's default is the factor 2 with no sweeps; a stream that outgrows its
burn-in 16-fold still rebuilds, and there the sweeps matter again.

\begin{table}[h]
\centering
\caption{Held-out log-likelihood (time) of streaming BLML under two re-binning tolerances; one row per
sample.}
\label{tab:rebin}
\begin{tabular}{rrccc}
\toprule
& & \multicolumn{2}{c}{tolerance $2^{1/4}$} & tolerance 2 \\
\cmidrule(lr){3-4}\cmidrule(lr){5-5}
$n$ & sample & no sweeps & two sweeps & (no rebuild) \\
\midrule
$2\cdot10^4$ & 0 & $-2.4429$ (16.5\,s) & $-2.4257$ (16.9\,s) & $-2.4265$ (15.7\,s) \\
$2\cdot10^4$ & 1 & $-2.4412$ (16.9\,s) & $-2.4327$ (17.5\,s) & $-2.4199$ (16.1\,s) \\
$2\cdot10^4$ & 2 & $-2.4150$ (17.2\,s) & $-2.4005$ (17.7\,s) & $-2.4005$ (15.8\,s) \\
$4\cdot10^4$ & 0 & $-2.2429$ (55.8\,s) & $-2.2262$ (59.8\,s) & $-2.2321$ (49.0\,s) \\
\bottomrule
\end{tabular}
\end{table}


\vskip 0.2in
\begin{thebibliography}{42}
\providecommand{\natexlab}[1]{#1}
\providecommand{\url}[1]{\texttt{#1}}
\expandafter\ifx\csname urlstyle\endcsname\relax
  \providecommand{\doi}[1]{doi: #1}\else
  \providecommand{\doi}{doi: \begingroup \urlstyle{rm}\Url}\fi

\bibitem[Abramson(1982)]{abramson1982}
Ian~S. Abramson.
\newblock On bandwidth variation in kernel estimates---a square root law.
\newblock \emph{The Annals of Statistics}, 10\penalty0 (4):\penalty0
  1217--1223, 1982.

\bibitem[Agarwal et~al.(2017)Agarwal, Chen, and Sarma]{agarwal2017blml}
Rahul Agarwal, Zhe Chen, and Sridevi~V. Sarma.
\newblock A novel nonparametric maximum likelihood estimator for probability
  density functions.
\newblock \emph{IEEE Transactions on Pattern Analysis and Machine
  Intelligence}, 39\penalty0 (7):\penalty0 1294--1308, 2017.
\newblock arXiv:1503.06236.

\bibitem[Bishop(1994)]{bishop1994mdn}
Christopher~M. Bishop.
\newblock Mixture density networks.
\newblock Technical Report NCRG/94/004, Aston University, 1994.

\bibitem[Bowman(1984)]{bowman1984}
Adrian~W. Bowman.
\newblock An alternative method of cross-validation for the smoothing of
  density estimates.
\newblock \emph{Biometrika}, 71\penalty0 (2):\penalty0 353--360, 1984.

\bibitem[Capp{\'e} and Moulines(2009)]{cappe2009online}
Olivier Capp{\'e} and Eric Moulines.
\newblock On-line expectation--maximization algorithm for latent data models.
\newblock \emph{Journal of the Royal Statistical Society: Series B},
  71\penalty0 (3):\penalty0 593--613, 2009.

\bibitem[Chen and Guestrin(2016)]{chen2016xgboost}
Tianqi Chen and Carlos Guestrin.
\newblock {XGBoost}: A scalable tree boosting system.
\newblock In \emph{Proceedings of the 22nd ACM SIGKDD International Conference
  on Knowledge Discovery and Data Mining}, pages 785--794, 2016.

\bibitem[Davis(1975)]{davis1975}
Kathryn~Bullock Davis.
\newblock Mean square error properties of density estimates.
\newblock \emph{The Annals of Statistics}, 3\penalty0 (4):\penalty0 1025--1030,
  1975.

\bibitem[Dawid(1984)]{dawid1984prequential}
A.~P. Dawid.
\newblock Present position and potential developments: Some personal views.
  statistical theory: The prequential approach.
\newblock \emph{Journal of the Royal Statistical Society, Series A},
  147\penalty0 (2):\penalty0 278--292, 1984.

\bibitem[Dinh et~al.(2017)Dinh, Sohl-Dickstein, and Bengio]{dinh2017realnvp}
Laurent Dinh, Jascha Sohl-Dickstein, and Samy Bengio.
\newblock Density estimation using {Real NVP}.
\newblock In \emph{International Conference on Learning Representations}, 2017.

\bibitem[Durkan et~al.(2019{\natexlab{a}})Durkan, Bekasov, Murray, and
  Papamakarios]{durkan2019neural}
Conor Durkan, Artur Bekasov, Iain Murray, and George Papamakarios.
\newblock Neural spline flows.
\newblock In \emph{Advances in Neural Information Processing Systems},
  volume~32, 2019{\natexlab{a}}.

\bibitem[Durkan et~al.(2019{\natexlab{b}})Durkan, Bekasov, Murray, and
  Papamakarios]{durkan2019nsf}
Conor Durkan, Artur Bekasov, Iain Murray, and George Papamakarios.
\newblock Neural spline flows.
\newblock In \emph{Advances in Neural Information Processing Systems},
  volume~32, 2019{\natexlab{b}}.

\bibitem[Germain et~al.(2015)Germain, Gregor, Murray, and
  Larochelle]{germain2015made}
Mathieu Germain, Karol Gregor, Iain Murray, and Hugo Larochelle.
\newblock {MADE}: Masked autoencoder for distribution estimation.
\newblock In \emph{Proceedings of the 32nd International Conference on Machine
  Learning}, pages 881--889, 2015.

\bibitem[Hall and Marron(1988)]{hall1988order}
Peter Hall and J.~S. Marron.
\newblock Choice of kernel order in density estimation.
\newblock \emph{The Annals of Statistics}, 16\penalty0 (1):\penalty0 161--173,
  1988.

\bibitem[Hollmann et~al.(2025)Hollmann, M{\"u}ller, Purucker, Krishnakumar,
  K{\"o}rfer, Hoo, Schirrmeister, and Hutter]{hollmann2025tabpfn}
Noah Hollmann, Samuel M{\"u}ller, Lennart Purucker, Arjun Krishnakumar, Max
  K{\"o}rfer, Shi~Bin Hoo, Robin~Tibor Schirrmeister, and Frank Hutter.
\newblock Accurate predictions on small data with a tabular foundation model.
\newblock \emph{Nature}, 637:\penalty0 319--326, 2025.

\bibitem[Hyndman et~al.(1996)Hyndman, Bashtannyk, and
  Grunwald]{hyndman1996conditional}
Rob~J. Hyndman, David~M. Bashtannyk, and Gary~K. Grunwald.
\newblock Estimating and visualizing conditional densities.
\newblock \emph{Journal of Computational and Graphical Statistics}, 5\penalty0
  (4):\penalty0 315--336, 1996.

\bibitem[Kingma and Ba(2015)]{kingma2015adam}
Diederik~P. Kingma and Jimmy Ba.
\newblock Adam: A method for stochastic optimization.
\newblock In \emph{International Conference on Learning Representations
  (ICLR)}, 2015.

\bibitem[Kristan et~al.(2011)Kristan, Leonardis, and
  Sko{\v{c}}aj]{kristan2011online}
Matej Kristan, Ale{\v{s}} Leonardis, and Danijel Sko{\v{c}}aj.
\newblock Multivariate online kernel density estimation with {G}aussian
  kernels.
\newblock \emph{Pattern Recognition}, 44\penalty0 (10--11):\penalty0
  2630--2642, 2011.

\bibitem[Lam et~al.(2015)Lam, Pitrou, and Seibert]{lam2015numba}
Siu~Kwan Lam, Antoine Pitrou, and Stanley Seibert.
\newblock Numba: A {LLVM}-based {P}ython {JIT} compiler.
\newblock In \emph{Proceedings of the Second Workshop on the LLVM Compiler
  Infrastructure in HPC}, 2015.

\bibitem[Landau and Pollak(1962)]{landau1962prolate}
H.~J. Landau and H.~O. Pollak.
\newblock Prolate spheroidal wave functions, {F}ourier analysis and
  uncertainty---{III}: The dimension of the space of essentially time- and
  band-limited signals.
\newblock \emph{Bell System Technical Journal}, 41\penalty0 (4):\penalty0
  1295--1336, 1962.

\bibitem[Landau and Widom(1980)]{landau1980eigenvalue}
H.~J. Landau and H.~Widom.
\newblock Eigenvalue distribution of time and frequency limiting.
\newblock \emph{Journal of Mathematical Analysis and Applications}, 77\penalty0
  (2):\penalty0 469--481, 1980.

\bibitem[Loftsgaarden and Quesenberry(1965)]{loftsgaarden1965}
D.~O. Loftsgaarden and C.~P. Quesenberry.
\newblock A nonparametric estimate of a multivariate density function.
\newblock \emph{The Annals of Mathematical Statistics}, 36\penalty0
  (3):\penalty0 1049--1051, 1965.

\bibitem[Maclaurin et~al.(2015)Maclaurin, Duvenaud, and
  Adams]{maclaurin2015hyper}
Dougal Maclaurin, David Duvenaud, and Ryan~P. Adams.
\newblock Gradient-based hyperparameter optimization through reversible
  learning.
\newblock In \emph{Proceedings of the 32nd International Conference on Machine
  Learning}, pages 2113--2122, 2015.

\bibitem[M{\"a}rz(2019)]{marz2019xgboostlss}
Alexander M{\"a}rz.
\newblock {XGBoostLSS} -- an extension of {XGBoost} to probabilistic
  forecasting.
\newblock \emph{arXiv preprint arXiv:1907.03178}, 2019.

\bibitem[Neal and Hinton(1998)]{neal1998em}
Radford~M. Neal and Geoffrey~E. Hinton.
\newblock A view of the {EM} algorithm that justifies incremental, sparse, and
  other variants.
\newblock In Michael~I. Jordan, editor, \emph{Learning in Graphical Models},
  pages 355--368. Kluwer, 1998.

\bibitem[Nesterov(2012)]{nesterov2012cd}
Yu. Nesterov.
\newblock Efficiency of coordinate descent methods on huge-scale optimization
  problems.
\newblock \emph{SIAM Journal on Optimization}, 22\penalty0 (2):\penalty0
  341--362, 2012.

\bibitem[Okuta et~al.(2017)Okuta, Unno, Nishino, Hido, and
  Loomis]{okuta2017cupy}
Ryosuke Okuta, Yuya Unno, Daisuke Nishino, Shohei Hido, and Crissman Loomis.
\newblock {CuPy}: A {NumPy}-compatible library for {NVIDIA} {GPU} calculations.
\newblock In \emph{Workshop on Machine Learning Systems (LearningSys),
  NeurIPS}, 2017.

\bibitem[Papamakarios et~al.(2017)Papamakarios, Pavlakou, and
  Murray]{papamakarios2017maf}
George Papamakarios, Theo Pavlakou, and Iain Murray.
\newblock Masked autoregressive flow for density estimation.
\newblock In \emph{Advances in Neural Information Processing Systems},
  volume~30, 2017.

\bibitem[Papamakarios et~al.(2021)Papamakarios, Nalisnick, Rezende, Mohamed,
  and Lakshminarayanan]{papamakarios2021normalizing}
George Papamakarios, Eric Nalisnick, Danilo~Jimenez Rezende, Shakir Mohamed,
  and Balaji Lakshminarayanan.
\newblock Normalizing flows for probabilistic modeling and inference.
\newblock \emph{Journal of Machine Learning Research}, 22\penalty0
  (57):\penalty0 1--64, 2021.

\bibitem[Pedregosa(2016)]{pedregosa2016hyper}
Fabian Pedregosa.
\newblock Hyperparameter optimization with approximate gradient.
\newblock In \emph{Proceedings of the 33rd International Conference on Machine
  Learning}, pages 737--746, 2016.

\bibitem[Rudemo(1982)]{rudemo1982}
Mats Rudemo.
\newblock Empirical choice of histograms and kernel density estimators.
\newblock \emph{Scandinavian Journal of Statistics}, 9\penalty0 (2):\penalty0
  65--78, 1982.

\bibitem[Scott(2015)]{scott2015}
David~W. Scott.
\newblock \emph{Multivariate Density Estimation: Theory, Practice, and
  Visualization}.
\newblock Wiley, 2nd edition, 2015.

\bibitem[Silverman(1982)]{silverman1982fft}
B.~W. Silverman.
\newblock Algorithm {AS} 176: Kernel density estimation using the fast
  {F}ourier transform.
\newblock \emph{Journal of the Royal Statistical Society, Series C (Applied
  Statistics)}, 31\penalty0 (1):\penalty0 93--99, 1982.

\bibitem[Silverman(1986)]{silverman1986}
Bernard~W. Silverman.
\newblock \emph{Density Estimation for Statistics and Data Analysis}.
\newblock Chapman and Hall, 1986.

\bibitem[Slepian(1976)]{slepian1976bandwidth}
D.~Slepian.
\newblock On bandwidth.
\newblock \emph{Proceedings of the IEEE}, 64\penalty0 (3):\penalty0 292--300,
  1976.

\bibitem[Tseng(2001)]{tseng2001}
Paul Tseng.
\newblock Convergence of a block coordinate descent method for
  nondifferentiable minimization.
\newblock \emph{Journal of Optimization Theory and Applications}, 109\penalty0
  (3):\penalty0 475--494, 2001.

\bibitem[Tsybakov(2009)]{tsybakov2009}
Alexandre~B. Tsybakov.
\newblock \emph{Introduction to Nonparametric Estimation}.
\newblock Springer, 2009.

\bibitem[Wand(1994)]{wand1994fast}
M.~P. Wand.
\newblock Fast computation of multivariate kernel estimators.
\newblock \emph{Journal of Computational and Graphical Statistics}, 3\penalty0
  (4):\penalty0 433--445, 1994.

\bibitem[Wand and Jones(1995)]{wand1995}
M.~P. Wand and M.~C. Jones.
\newblock \emph{Kernel Smoothing}.
\newblock Chapman and Hall, 1995.

\bibitem[Wegman and Davies(1979)]{wegman1979recursive}
Edward~J. Wegman and H.~I. Davies.
\newblock Remarks on some recursive estimators of a probability density.
\newblock \emph{The Annals of Statistics}, 7\penalty0 (2):\penalty0 316--327,
  1979.

\bibitem[Wolverton and Wagner(1969)]{wolverton1969}
C.~T. Wolverton and T.~J. Wagner.
\newblock Asymptotically optimal discriminant functions for pattern
  classification.
\newblock \emph{IEEE Transactions on Information Theory}, 15\penalty0
  (2):\penalty0 258--265, 1969.

\bibitem[Wright(2015)]{wright2015cd}
Stephen~J. Wright.
\newblock Coordinate descent algorithms.
\newblock \emph{Mathematical Programming}, 151\penalty0 (1):\penalty0 3--34,
  2015.

\bibitem[Yao(1967)]{yao1967}
Kung Yao.
\newblock Applications of reproducing kernel {H}ilbert spaces---bandlimited
  signal models.
\newblock \emph{Information and Control}, 11\penalty0 (4):\penalty0 429--444,
  1967.

\end{thebibliography}
\end{document}